\documentclass[11pt]{book}

\usepackage{shortex}
\usepackage[margin=1.5in]{geometry}
\usepackage{algorithm}
\usepackage{algpseudocode}
\usepackage[numbers,sort&compress]{natbib}

\parsefontstylestrings{{c}}{\uppercaseRoman}
\parsefontstylestrings{{c}}{\lowercaseRoman}
\DeclareMathOperator{\MH}{MH}

\makeatletter
\newcommand{\chapterauthor}[1]{%
	{\parindent0pt\vspace*{-25pt}%
		\linespread{1.1}\large\scshape#1%
		\par\nobreak\vspace*{35pt}}
	\@afterheading%
}
\makeatother

\begin{document}

\chapter*{MCMC-driven learning}
\chapterauthor{Alexandre Bouchard-C\^ot\'e (\href{mailto:bouchard@stat.ubc.ca}{bouchard@stat.ubc.ca}) \\ 
Trevor Campbell (\href{mailto:trevor@stat.ubc.ca}{trevor@stat.ubc.ca}) \\
Geoff Pleiss (\href{mailto:geoff.pleiss@stat.ubc.ca}{geoff.pleiss@stat.ubc.ca}) \\ 
Nikola Surjanovic (\href{mailto:nikola.surjanovic@stat.ubc.ca}{nikola.surjanovic@stat.ubc.ca})}
\chaptermark{MCMC-driven learning}

\section{Introduction}
\label{sec:intro}

Markov chain Monte Carlo (MCMC) is a general computational methodology for
obtaining approximate draws from a target probability distribution $\pi$ by
simulating a Markov chain with stationary distribution $\pi$.  Historically,
MCMC has been used mostly in situations with a \emph{fixed} target distribution
$\pi$ that admits pointwise density evaluation up to a constant; this is the
usual case for Bayesian posterior inference, which is one of the most fruitful application
areas for MCMC. Similarly, the majority of MCMC research in computational
statistics has used a \emph{fixed} Markov kernel $\kappa$.

In the last two decades, MCMC has been increasingly used in situations where
either the target $\pi$, the kernel $\kappa$, or both involve parameters $\phi
\in \Phi \subset \reals^m$ that are tuned based on the generated Markov chain
sequence. Originally this research area was driven by work on adaptive MCMC
methods \cite{andrieu2008tutorial} whose goal was still to take draws from a fixed target $\pi$,
but with an adaptive kernel $\kappa_\phi$ in the interest of improved mixing.
Adaptivity arose from either varying scale/step size parameters in traditional
kernels \cite{haario_adaptive_2001,andrieu_ergodicity_2006,roberts2009examples}, by adding parametrized families to
generate independent proposals \cite{marzouk2016introduction}, or as a
surrogate for fast proposal generation \cite{rasmussen2003gaussian}.
Connections to the stochastic approximation literature (first noted by Andrieu \& Robert \cite{andrieu_controlled_mcmc_2001}),
and more recently advances from machine learning (ML), have created the need to study
and develop MCMC methods in situations where both the kernel $\kappa_\phi$ and
target $\pi_\phi$ may be adaptive, and the goal is not just to produce samples
from $\pi_\phi$ but perhaps also to minimize some cost associated with $\phi$---often
a discrepancy between $\pi_\phi$ and some other target $\pi$.

The goal of this chapter is to unify all of these problems---which encompass
black-box variational inference 
\cite{blei2017variational,kucukelbir2017automatic,ranganath2014black,zhang2018advances},
adaptive MCMC \cite{andrieu2008tutorial},
normalizing flow construction and transport-assisted MCMC 
\cite{arbel2021aft,gabrie2022adaptive, hagemann2022snf, matthews2022craft, samsonov2022local, wu2020snf},
surrogate-likelihood MCMC \cite{jarvenpaa2019efficient,meeds2014gps,wood2010statistical},
coreset construction for MCMC with big data 
\cite{campbell2019sparse,campbell2018bayesian,campbell2019automated,chen2023coreset,
chen2022bayesian,huggins2016coresets,manousakas2020bayesian,naik2022fast},
Markov chain gradient descent \cite{sun2018mcgd}, 
Markovian score climbing \cite{kim2022markov,naesseth2020markovian}, and more; 
see \cref{tab:moi_summary}---within one common framework, so that theory and
methods developed for each may be translated and generalized. To unify this
wide range of seemingly different MCMC/ML tasks, we introduce the notion of a
\emph{Markovian optimization-integration (MOI) problem},\footnote{MOI problems 
have been studied extensively in the field of \emph{stochastic approximation} \cite{benveniste2012adaptive}. 
Problems of the form \cref{eq:key}, but with different assumptions on estimates
of the expectation (e.g., unbiasedness), are sometimes called stochastic approximation problems \cite[Section 2]{moulines_non-asymptotic_2011}. 
In the Markovian setting, estimates are biased due to non-stationarity; we use the term MOI to avoid confusion, 
but note that \cref{eq:key} does have a rich antecedent in the literature \cite{andrieu_ergodicity_2006,benveniste2012adaptive}.} which involves
finding $\phi \in \Phi \subset \reals^m$ such that 
\[
  \label{eq:key} 
  g(\phi) := \ex_{\pi_\phi}\lt[g(X,\phi)\rt] = 0,
\]
where we do not have access to \iid draws $X\in\fcX$ from $\pi_\phi$, but rather
for each $\phi \in \Phi$ we are able to simulate from a $\pi_\phi$-invariant Markov chain with transition kernel 
$\kappa_\phi$ \cite{benveniste2012adaptive}.\footnote{More generally, we may 
assume that we can simulate from a Markov kernel 
that has an augmented stationary distribution that admits $\pi_\phi$ as a marginal.}
We are often motivated by cases where $g : \fcX \times \Phi \to \reals^{m'}$ in  
\cref{eq:key} is the gradient of a function that we seek to optimize, i.e., 
$g(x,\phi) = -\nabla_\phi f(x,\phi)$ for an objective $f$,
and we are interested in solving
\[ 
  \label{eq:min}
  \min_\phi f(\phi). 
  \qquad f(\phi):= \ex_{\pi_\phi} \lt[f(X, \phi)\rt].
\]
As we illustrate in this chapter, MOI problems of the form \cref{eq:key,eq:min} capture a wide
range of both well-established and cutting-edge research areas 
at the intersection of MCMC and ML.  For example, standard Bayesian
\emph{variational inference}---i.e., minimization of the reverse KL divergence
$\kl{q_\phi}{\pi} = \int \log\frac{q_\phi(x)}{\pi(x)} q_\phi(x) \, \dee x$ 
within a family $q_\phi \in \fcQ$ that admits \iid
draws---can be seen as an MOI problem by setting $f(\phi) = \kl{q_\phi}{\pi}$
and $\pi_\phi = q_\phi$ in \cref{eq:min}.  But so can more challenging
settings, such as minimization of the reverse KL divergence when $q_\phi$ does
not admit \iid draws (as in coreset construction \cite{chen2023coreset}), or minimization of
the forward KL divergence, by setting $f(\phi) = \kl{\pi}{q_\phi}$ and
$\pi_\phi = \pi$ in \cref{eq:min} 
\cite{kim2022markov,naesseth2020markovian,surjanovic2022parallel}.

By unifying these problems under the MOI framework, we see commonalities
among the optimization procedures used to solve them.
Existing algorithms designed for specific MOI problems can often be cast as
variants of the adaptive algorithm
\[
  \label{eq:rm}
  \phi_{t+1} &= \phi_t + \gamma_t g(X_t, \phi_t)  \\
  X_{t+1} &\dist \kappa_{\phi_{t+1}}(\cdot | X_t),
\]
where $\gamma_t$ is the step size at iteration $t \geq 0$
and $\kappa_\phi$ is a Markov kernel that targets $\pi_\phi$.
This algorithm, and in particular the convergence of the parameters $\phi_t$,
has been studied extensively for the general setup in \cref{eq:key} 
\cite{andrieu_stability_2005,andrieu_ergodicity_2006,andrieu_expanding_projections_2014,benveniste2012adaptive}.
But this chapter also highlights deviations from this general formula,
both in terms of the algorithmic approach \cite{surjanovic2022parallel} and the theoretical
analysis techniques used to demonstrate convergence \cite{chen2023coreset} for
specific instances of \cref{eq:key}. 

The chapter concludes with a detailed exploration of a case of particular interest: 
distribution approximation via minimization of the forward KL divergence $f(\phi) = \kl{\pi}{q_\phi}$.
Traditionally, the forward KL has not been considered in divergence minimization problems 
due to the inavailability of \iid draws from $\pi$, making the minimization of this 
divergence generally more difficult. 
However, the forward KL is of particular interest because it has the property 
that it is mass covering; for instance, 
in the case of a multimodal distribution $\pi$, an optimal approximating distribution $q_{\phi^\star}$
under the forward KL would generally avoid undesirable mode collapse.
Our example first considers training $q_\phi$ for use within an independence Metropolis--Hastings (IMH)
proposal. We then generalize IMH to the case of variational parallel tempering 
\cite{surjanovic2022parallel}, which targets distributions along an annealing path 
$q_\phi^{1-\beta} \cdot \pi^\beta$ for $0 \leq \beta \leq 1$; here the forward KL
is particularly desirable to minimize as it is known to control 
performance \cite{surjanovic2022parallel,syed2021parallel}.

This chapter is organized as follows. 
In \cref{sec:examples}, we list important examples of MOI problems, 
mapping them onto \cref{eq:key,eq:min}.
Next, in \cref{sec:strategies}, we cover general strategies for solving MOI problems in greater detail, 
and in \cref{sec:theory} we present theoretical results on stability and convergence.
Because the MOI framework unifies many different algorithms at the intersection of 
MCMC and ML, the accompanying theory can therefore be applied to a wide range 
of problems.
Finally, in \cref{sec:dist_approx}, we work through a case study
introducing several ideas and highlighting techniques previously presented in the chapter.
We end with a discussion in \cref{sec:discussion} and a summary of suggested further 
reading in \cref{sec:further_reading}.

\begin{table}
  \centering
  \caption{
    Common MOI problems and a classification of the primary aspect of interest: 
    either samples from $\pi$ (sampling problem) and/or an approximation 
    $q_\phi$ of $\pi$ (optimization problem). 
    Optional settings are indicated by $\sim$.
    $\star$ denotes a problem described in detail in \cref{sec:dist_approx}.
    \label{tab:moi_summary}
  }
  \resizebox{\linewidth}{!}{%
    \begin{tabular}{cccccc}
      \toprule
        Problem &&
        \shortstack{Sampling \\ problem?} &
        \shortstack{Optimization \\ problem?} &
        \shortstack{Adaptive \\ $\kappa_\phi$?} &
        \shortstack{Adaptive \\ $\pi_\phi$?}
        \\
      \midrule
        Black-box VI \cite{blei2017variational,kucukelbir2017automatic,ranganath2014black,zhang2018advances} &&
        & \checkmark & \checkmark & \checkmark \\
        Forward KL VI/Score climbing\textsuperscript{$\star$} \cite{kim2022markov,naesseth2020markovian} &&
        & \checkmark & $\sim$ & $\sim$ \\
        Adaptive MCMC \cite{andrieu2008tutorial} &&
        \checkmark & & \checkmark & \\
        Transport-assisted MCMC \cite{arbel2021aft,gabrie2022adaptive, hagemann2022snf,
          matthews2022craft, samsonov2022local, wu2020snf} &&
        \checkmark & \checkmark & $\sim$ & $\sim$ \\
        Surrogate-likelihood MCMC \cite{jarvenpaa2019efficient} &&
        \checkmark & & \checkmark & \checkmark \\
        Coreset MCMC
          \cite{campbell2019sparse,campbell2018bayesian,campbell2019automated,chen2023coreset,
          chen2022bayesian,huggins2016coresets,manousakas2020bayesian,naik2022fast} &&
        \checkmark & \checkmark & \checkmark & \checkmark \\
        Markov chain gradient descent \cite{sun2018mcgd} &&
        & \checkmark &  &  \\ %
        Variational parallel tempering\textsuperscript{$\star$} \cite{surjanovic2022parallel,syed2021parallel} &&
        \checkmark & \checkmark & \checkmark & \checkmark \\
      \bottomrule
    \end{tabular}
  }
  \label{tab:common_MOI}
\end{table}

\section{Examples of MOI problems}
\label{sec:examples}

We give an overview of various problems that have the MOI mathematical
structure, given by \cref{eq:key} and \cref{eq:min},
in order to motivate and emphasize the prevalence of such problems.
A summary is provided in \cref{tab:common_MOI}.

\subsubsection{Forward KL variational inference}
Bayesian variational inference (see, e.g., \cite{blei2017variational,kucukelbir2017automatic,ranganath2014black,zhang2018advances})
targeting a wide range of divergences is a common example of an MOI problem.
One divergence of particular interest in statistical applications is the forward (i.e., inclusive) KL divergence:
\[
  \min_\phi \kl{p}{q_\phi} &= \min_\phi \int p(x) \log\frac{p(x)}{q_\phi(x)} \, \dee x,
\]
We can view this problem as an MOI problem by setting
$\pi_\phi = p$ and $f(x, \phi) = \log\frac{p(x)}{q_\phi(x)}$, where 
$\kappa_\phi$ is a Markov chain targeting $p$.
This framework is quite general, encompassing techniques like
Markovian score climbing \cite{kim2022markov,naesseth2020markovian} and  cases
where $q_\phi$ does not have a tractable normalizing constant.

In \cref{sec:dist_approx},
we work through the derivation of a forward KL minimization algorithm
where the variational family contains the prior $\pi_0$
and the MCMC samples are obtained on a transformed space.
Specifically, given a tractable distribution $f_0$ and function
$G$ such that $f_0 \circ G^{-1} = \pi_0$, we set the target distribution
on the transformed space to be $\tilde\pi(z) \propto f_0(z) \; L(G(z)|x)$, 
where $L(\cdot | x)$ is the likelihood.
If $f_0$ lies withing some variational family $\cbra{f_\phi : \phi \in \Phi}$,
we consider $\min_\phi \kl{\tilde \pi}{f_\phi}$, instead of $\min_\phi \kl{\pi}{q_\phi}$.
That is, we set $\pi_\phi = \tilde \pi$, $\kappa_\phi$ is a Markov kernel targeting
$\tilde \pi$, and $f(x, \phi) = \log \frac{\tilde\pi(x)}{f_\phi(x)}$.

Another perhaps less obvious instantiation of forward KL minimization is
parallel tempering with a variational reference $q_\phi$
\cite{surjanovic2022parallel}.  
We
provide a detailed discussion of this problem in \cref{sec:variational_pt}.

\subsubsection{Reverse KL variational inference}
The most common divergence to use in Bayesian variational inference is the reverse (i.e., exclusive) KL divergence:
\[
  \min_\phi \kl{q_\phi}{p} &= \min_\phi \int q_\phi(x) \log\frac{q_\phi(x)}{p(x)} \, \dee x.
\]
Here $p$ is the target distribution/density of interest and $q_\phi$ belongs to a family of
distributions parameterized by $\phi \in \Phi$ that each enables tractable \iid draws and pointwise density evaluation (although
see the coreset MCMC example below for a reverse KL minimization problem without the availability of \iid draws or density evaluation).
In the setup of \cref{eq:min}, we have 
$\pi_\phi = q_\phi$, $\kappa_\phi(\dee x | x) = q_\phi(\dee x)$, and 
$f(x, \phi) = \log\frac{q_\phi(x)}{p(x)}$; note here that since $q_\phi$ enables \iid draws, the kernel $\kappa_\phi$
does not depend on the previous state.
This optimization problem encompasses a wide range of approaches to inference, including methods
such as normalizing flows \cite{tabak_flows_2010,tabak_flows_2013,rezende_flows_2015,papamakarios2021normalizing},
amortized variational inference \cite{kingma2014auto}, and variational annealing families \cite{Geffner21,Zhang21,Jankowiak21}, 
among others.

There are various extensions and modifications to this setup. For example, if the target $p = p_\phi$ has its own parameters
(but is still known up to a constant, e.g., when $p_\phi \propto p_\phi(z) p_\phi(x|z)$ with tractable prior/likelihood families),
then this approach becomes variational expectation--maximization.
Another common modification is a reparameterization of
the problem above, resulting in an objective of the form
\[
  \min_\phi \kl{q_\phi}{p} &= \min_\phi \int q(x) \log\frac{q(x)}{p_\phi(x)} \, \dee x,
\]
where $p_\phi$ is the transformed version of $p$, $\kappa_\phi(\dee x | x) = q(\dee x)$, 
and $f(x, \phi) = \log\frac{q(x)}{p_\phi(x)}$. Even after reparameterization, this 
problem is still expressable within the MOI framework.

\subsubsection{Adaptive MCMC}

In the adaptive MCMC literature (see, e.g., \cite{andrieu2008tutorial}),
we are typically provided a parametrized kernel $\kappa_\phi$ that targets a fixed distribution $\pi$, 
and an objective function that guides $\phi$ towards a value that approximately minimizes the autocorrelation
of the Markov chain with kernel $\kappa_\phi$. 
For instance, a common metric is the expected squared jump distance at stationarity. 
Let $X_1 \sim \pi$ and $X_2 | X_1 \sim \kappa_\phi(X_1, \cdot)$. 
Then, our objective may be
\[
  \min_\phi -\ex\lt[ (X_1 - X_2)^2 \rt].
\]
To express this in the MOI framework, we note that the joint distribution of $X = (X_1, X_2)$ 
is $\pi \otimes \kappa_\phi$. 
Then, $\pi_\phi = \pi \otimes \kappa_\phi$, and $f(x, \phi) = -(x_1 - x_2)^2$, where $x = (x_1, x_2)$.
Alternatively, one might target a particular (theoretically optimal) 
acceptance rate $\alpha^\star \in [0,1]$,
\[
  \min_\phi \ex\lt[ (\alpha(X_1,X_2) - \alpha^\star)^2 \rt],
\]
where $\alpha(x_1, x_2)$ is a valid MH acceptance probability.
This approach can be useful when a theoretically optimal choice of acceptance rate $\alpha^\star$ is known, 
such as for random-walk MH \cite{roberts1997weak} and the Metropolis-adjusted Langevin algorithm 
\cite{roberts1998optimal}. Here, the setup is the same as above, except that we take 
$f(x, \phi) = (\alpha(x_1, x_2) - \alpha^\star)^2$.
Alternative objectives for adaptive MCMC can be incorporated into the MOI framework similarly.

\subsubsection{Transport-assisted MCMC}

Transport assisted MCMC methods build invertible transport maps $f_\phi$ such that
$q_\phi = q \circ f_\phi^{-1} \approx \pi$, where $q$ is some simple distribution 
such as a standard multivariate normal.
We provide a more detailed overview of these methods in \cref{sec:transport};
here we briefly summarize the two high-level approaches for using transport maps.
The first approach uses MCMC to obtain approximate samples $X_1, \ldots, X_t$
from $\pi \circ f_\phi$, which should have a simpler geometry. These samples are then 
transported to $f_\phi(X_1), \ldots, f_\phi(X_t)$, yielding an approximation to $\pi$. 
The second approach uses $q_\phi$ as a proposal within IMH (or some variant thereof)
and learning the appropriate parameters $\phi$. 
In both cases, learning the variational parameters $\phi$ can often be subsumed 
into the forward or reverse KL variational inference framework above. For instance, 
in the second approach to transport-assisted MCMC with a Markov kernel targeting $\pi$, 
one can compose an IMH proposal from $q_\phi$ with 
any other kernel $\kappa$ targeting $\pi$ \cite{gabrie2022adaptive}. 
In this case, one may hope that an appropriate proposal $q_\phi$ is learned to 
allow for large global moves, while the Markov kernel $\kappa$ explores locally.

\subsubsection{Surrogate-based inference}

In many large-scale applications, we may wish to sample from a posterior
$\pi(x) \propto L(\mathcal D \mid x) \pi_0(x)$,
where the likelihood $L(\cdot \mid x)$ can only be accessed through sampling.
Approximate Bayesian Computaition (ABC) algorithms \cite{beaumont2002approximate,rubin1984bayesianly}
aim to produce an approximate posterior
based on limited and potentially noisy samples $\mathcal D' \sim L(\cdot \mid x)$.
A recent trend in the ABC literature is to
 approximate $L(\mathcal D \mid x)$ with a tractable
surrogate density $q_\phi(\cdot \mid x)$ which can be used in conjunction with
MCMC or approximate inference methods to produce approximate posterior samples
drawn from $\pi_\phi(x) \propto q_\phi(\mathcal D \mid x) \pi_0(x)$.
In its simplest form, $q_\phi(\cdot \mid x)$ is a simple parametric density
(e.g., a multivariate normal) where parameters are estimated from
repeated evaluations of $L(\cdot \mid x)$ \cite{price2018bayesian,wood2010statistical}.
Recent work uses more powerful density estimates for
$q_\phi(\cdot \mid x)$ such as normalizing flows \cite{papamakarios2019sequential}.
More complex instantiations approximate $q_\phi(\cdot \mid x)$
with a probabilistic model, such as a Gaussian process
\cite{meeds2014gps,wilkinson2014accelerating},
and uncertainty over $q_\phi(\cdot \mid x)$ is used in conjunction with
the MCMC kernel to determine where to evaluate $L$
\cite{gutmann2016bayesian,jarvenpaa2019efficient,kandasamy2017query}.

Surrogate-based inference algorithms can often be cast as MOI problems.
For example, \cite{papamakarios2019sequential} propose the following
adaptive loop for optimizing the surrogate likelihood $q_\phi(\cdot \mid x)$:
\begin{enumerate}
  \item $x_i$ is sampled from $\pi_{\phi_{i-1}}(\cdot) \propto q_{\phi_{i-1}}(\mathcal
    D \mid \cdot) \pi_0(\cdot)$ via MCMC,
  \item $\mathcal D'_i$ is sampled from $L(\cdot | x_i)$, and
  \item the surrogate likelihood parameters are updated via maximum likelihood:
    \[
      \label{eq:surrogate_update}
      \phi_i = \argmin_\phi \tfrac{1}{i} \sum_{j=1}^i
      \log q_{\phi_{i-1}} \left( \mathcal D'_j \mid x_j \right).
    \]
\end{enumerate}
Note that the objective in \cref{eq:surrogate_update} is an unbiased estimate
of the summation of cross entropies
$\tfrac{1}{i} \sum_{j=1}^i \ex_{L(\mathcal D' \mid x_j)}
[ \log q_{\phi_{i-1}} ( \mathcal D' \mid x_j ) ].$
Assuming that \cref{eq:surrogate_update} is optimized by gradient descent,
this iterative procedure can be viewed as an instantiation of \cref{eq:rm}
and thus optimizes the following MOI problem:
\[
  \min_\phi \ex_{\pi_\phi(x)} \left[
    \ex_{L(\mathcal D' \mid x)} \left[
      \log \left( q_\phi( \mathcal D' \mid x_j \right)
    \right]
  \right].
\]

\subsubsection{Coreset MCMC}\label{subsec:coresetexample}

When the size of a dataset is large, traditional MCMC algorithms can suffer due to the 
need to examine the entire dataset for each (gradient) log-likelihood computation.
With $N$ data points, this evaluation takes $O(N)$ time, and hence is infeasible 
when $N$ is on the order of millions or billions of data points.
One way around this issue is to construct a representative subsample of 
the data, referred to as a \emph{coreset} 
\cite{campbell2019sparse,campbell2018bayesian,campbell2019automated,chen2023coreset,
chen2022bayesian,huggins2016coresets,manousakas2020bayesian,naik2022fast}.
In certain settings, there exists a coreset of size $O(\log N)$ that suffices to summarize a dataset of size $N$,
resulting in an exponential reduction in computational cost \cite{chen2022bayesian,naik2022fast};
the difficulty remains in accurately and efficiently finding such a coreset.
Formally, if we express 
\[
  \pi(x) \propto \pi_0(x) \cdot \exp\lt(\sum_{n=1}^N \ell_n(x)\rt),
\]
the goal of coreset construction is to find $M \ll N$ data points and corresponding 
weights $\cbra{w_m}_{m=1}^M$ such that the coreset approximation to the posterior, $\pi_w$, is 
close to $\pi$. In expressing $\pi_w$, we assume without loss of generality that 
the subsample consists of the first $M$ data points, so that 
\[
  \pi_w(x) \propto \pi_0(x) \cdot \exp\lt(\sum_{m=1}^M w_m \ell_m(x)\rt).
\]

Some coreset construction techniques fall within the MOI framework. 
For instance, following the approach of \cite{chen2023coreset}, the coreset
construction problem is posed as a reverse KL divergence minimization problem:
\[
  \min_{w \in \fcW} \kl{\pi_w}{\pi}, \qquad  
  \fcW = \cbra{w \in \reals^M : w_m \geq 0}.
\]
Hence, we can set $\phi = w$, $\pi_\phi = \pi_w$, and $f(x,\phi) = \log\frac{\pi_w(x)}{\pi(x)}$. 
In this case, since $\pi_w$ does not enable \iid draws, we are forced to use a kernel
$\kappa_\phi$ targeting $\pi_\phi = \pi_w$ to obtain estimates of the divergence gradient \cite{chen2023coreset}.
\subsubsection{Markov chain gradient descent}

Consider minimization of a function $f:\Phi \to \reals$ that can be expressed as 
\[ \label{eq:mcgd}
  f(\phi) = \ex_\pi[f(X,\phi)],   
\]
for some $f: \fcX \times \Phi \to \reals$ and a distribution $\pi$ on $\fcX$. 
For instance, if $f(\phi) = N^{-1} \sum_{n=1}^N f_n(\phi)$, then it can be expressed 
in the form \cref{eq:mcgd}. 
In stochastic gradient descent one would draw $X \sim \pi$ and use $f(X,\phi)$ as an 
estimate of $f(\phi)$ to perform a gradient update. 
In contrast, Markov chain gradient descent methods \cite{sun2018mcgd} use a fixed Markov kernel $\kappa$ that 
admits $\pi$ as a stationary distribution to estimate $f(\phi)$. 
In the MOI framework, this corresponds to setting $\pi_\phi = \pi$ and $\kappa_\phi = \kappa$.
This approach can be useful when direct simulation of $\pi$ is difficult. 
Another example where such an approach might be useful is if the data points 
$f_n(x)$ are stored across multiple machines, as highlighted by \cite{sun2018mcgd}, 
in which case it may be desirable to form a random walk along data points that minimizes communication 
costs between machines.

\section{Strategies for MOI problems}
\label{sec:strategies} 

In this section we offer some wisdom for solving various MOI problems.
Our review touches upon the reparameterization trick and REINFORCE techniques for 
gradient estimation, automatic differentitation, other gradient descent techniques, 
and methods for stabilizing gradient estimates.

\subsection{Stochastic gradient estimation}

Consider the MOI optimization problem \cref{eq:min}.
Classical methods rely on the exact computation of $\nabla f$. However, due to the intractable
expectation in \cref{eq:min}, this is not generally possible in the MOI context. 
Instead, many methods such as stochastic gradient descent rely on an 
\emph{estimate} of $\nabla f$. In other words, a random variable $\hat G$ such that 
$\hat G = g(X_t, \phi_t)$ approximates $-\nabla f(\phi_t)$. 
In this section we review methods to construct such approximations. 

First, we clarify what is meant by the assertion that ``$\hat G$ approximates $\nabla f$.''
In a large segment of the stochastic optimization literature, 
the hypothesis on $\hat G$ is that it is an unbiased estimate, namely that 
$\ex[\hat G | X_1, \dots, X_{t-1}] = - \nabla f(\phi_t)$ (see e.g. \cite[Section 2]{moulines_non-asymptotic_2011}).
In our context, unbiasedness only holds in the special case
where  $X_t \sim \pi_{\phi_t}$, but 
recall that the MOI setup instead allows for the more general 
situation where $X_t \sim \kappa_{\phi_{t}}(\cdot | X_{t-1})$, with $\kappa_{\phi_t}$ 
being a $\pi_{\phi_t}$-invariant Markov kernel. 
As a result, the gradient estimators considered for MOI problems often
do not satisfy the unbiasedness condition. 
As covered later in \cref{sec:theory}, the theory in this more general Markovian setting is rather technical.
As a result, the MOI methodological literature has so far generally used gradient estimators 
$\hat G$ that originate from the unbiasedness setting, but where the Markovian $X_t$ 
is plugged-in instead of the \iid counterpart. 

In the remainder of this section we start by covering the two most 
popular estimators in machine learning: one based on the reparameterization 
trick (for the \iid special case), and the REINFORCE estimator. 
For a more in-depth review of stochastic gradient estimation, 
we refer readers to \cite{mohamed_monte_2020}.  
Both the reparameterization trick and the REINFORCE estimator reduce the problem 
of computing the gradient of an expectation 
to the problem of computing the gradient of a function. 
However, the function to be differentiated might be complicated, and so the success of these 
methods often depends on automatic differentiation methods, 
reviewed in \cref{sec:autodiff}.

Before going into further details on gradient estimators, 
we emphasize a point that we later expand on in \cref{sec:dist_approx}: in the MOI context 
it is sometimes possible to avoid gradient estimation altogether and instead 
transform the optimization problem into an ``auxiliary statistical estimation problem.''
However, in other situations, gradient estimation might be unavoidable.

\subsubsection{Reparameterization trick}

We review a gradient estimator first introduced in the 
operations research literature under the name of the ``push-out'' 
estimator \cite{rubinstein_sensitivity_1992}, and later reinvented in the 
machine learning literature under the name of the 
``reparameterization trick'' \cite{kingma2014auto,rezende2014stochastic}.
We use the more popular machine learning term for clarity.

When performing gradient-based MOI optimization, one is usually required to compute 
gradients of the form $\nabla_\phi \ex_{q_\phi}[f(X,\phi)]$, for some function $f$. 
For instance, $f$ may be the difference of log densities $f(x,\phi) = \log\frac{q_\phi(x)}{\pi(x)}$, 
in which case the gradient can be used for minimization of the reverse KL 
divergence $\kl{q_\phi}{\pi}$.

Suppose that there exists a tractable base distribution $q$ and a map $m_\phi$ 
such that $q_\phi = q \circ m_\phi^{-1}$. The reparameterization trick is based 
on the observation that 
\[
  \nabla_\phi \ex_{q_\phi}[f(X,\phi)] 
  &= \nabla_\phi \ex_{q}[f(m_\phi(X),\phi)] 
  = \int \nabla_\phi f(m_\phi(x),\phi) q(x) \, \dee x,
\]
provided that $f$ is differentiable with respect to both arguments, that 
$m_\phi$ is differentiable with respect to $\phi$, and the existence of 
an integrable envelope for $\nabla_\phi f(m_\phi(x),\phi)$.
An unbiased estimate of this gradient can be obtained with
\[
  \hat\nabla_\phi \ex_{q_\phi}[f(X,\phi)] 
  = T^{-1} \sum_{t=1}^T \nabla_\phi f(m_\phi(X_t),\phi), 
\]
where $\cbra{X_t}_{t=1}^T$ are \iid samples from $q$.

Although the reparameterization trick is a common way of obtaining estimates of such gradients,
it is important to keep in mind that it only applies in certain settings.
For instance, the restrictive condition on the differentiability of $f$ with respect 
to both arguments needs to be satisfied.
In the next section, we introduce a technique called REINFORCE that is applicable 
in a wider range of settings, albeit at the expense of potentially higher-variance 
gradient estimates.

\subsubsection{REINFORCE}

In settings where the reparameterization trick presented above cannot be applied, such
as when the derivatives of $f(x,\phi)$ with respect to both $x$ and $\phi$ 
are not available, the REINFORCE method may be applicable.\footnote{Again, the REINFORCE terminology is the more 
popular machine learning term, but the method was discovered earlier 
under the name ``likelihood ratio method'' \cite{glynn_likelihood_1990}.}
Generally, REINFORCE estimates of gradients have been observed to have higher 
variance than corresponding estimates based on the reparameterization trick.
Central to this method is the identity that 
\[
  \nabla_\phi q_\phi = q_\phi \nabla_\phi \log q_\phi,
\]
combined with an interchange of an integral and a derivative under appropriate assumptions. 

Consider the same gradient estimation problem as above for the reparameterization trick. 
Using the identity above, note that 
\[
  \nabla_\phi \ex_{q_\phi}[f(X,\phi)] 
  &= \nabla_\phi \int f(x,\phi) q_\phi(x) \, \dee x \\
  &= \int \cbra{\nabla_\phi f(x,\phi) q_\phi(x) + f(x,\phi) \nabla_\phi q_\phi(x)} \, \dee x \\
  &= \int \cbra{\nabla_\phi f(x,\phi) q_\phi(x) + f(x,\phi) \nabla_\phi \log q_\phi(x) q_\phi(x)} \, \dee x,
\]
where this time the only differentiability of $f$ is with respect to its second argument $\phi$ 
(combined with an integrable envelope condition, this time on $\nabla f(x,\phi) q_\phi(x)$). 
This gradient can therefore be approximated with 
\[
  \hat\nabla_\phi \ex_{q_\phi}[f(X,\phi)] 
  = T^{-1} \sum_{t=1}^T \cbra{\nabla_\phi f(X_t, \phi) + f(X_t, \phi) \nabla_\phi \log q_\phi(X_t)},
\]
where $\cbra{X_t}_{t=1}^T$ are \iid samples from $q_\phi$.

A drawback of the REINFORCE method is the empirical observation that is often (but not always)
have high variance compared to the reparametrization trick \cite{mohamed_monte_2020}. 
To counteract this inflated variance, it is common to use \emph{control variates}, 
which is a technique that we present in \cref{sec:gradient_stabilization}.

\subsection{Automatic differentation}
\label{sec:autodiff}

The methods presented in the preceding section all involve computation of a gradient of 
a function $f$, which may potentially be very complex. 
For instance, $f$ may be a neural network with a complicated architecture. 
With this in mind, how can one go about implementing various MOI problems without painstakingly 
deriving expressions for the gradients each time?

\emph{Automatic differentiation} (AD) \cite{morgenstern_how_1985}  provides an
elegant solution to this problem. 
AD takes as input the code for a function and outputs a new function computing the gradient. 
 It bypasses the issues of symbolic computation of gradients 
and numerically-unstable finite differencing techniques.
As a now widespread approach to differentiation \cite{baydin2018autodiff}, AD combines the worlds of symbolic and 
numeric differentiation and is superior to either alone in the context of 
solving MOI problems: gradients are computed using symbolic differentiation rules 
but are represented numerically (as augmented dual numbers) in the machine. 
In essence, the computer program itself is differentiated, resulting in a 
programming paradigm referred to as \emph{differentiable programming}.

AD has proven to be an indispensable tool for solving MOI problems and has revolutionized 
the field of machine learning, allowing users to implement complicated architectures 
without worrying about the details of gradient calculation for the purpose of 
optimizing hyperparameters.
The AD approach to differentiation is 
fundamentally different from traditional approaches such as 
symbolic and numerical differentiation and bypasses issues presented with both. 
For instance, numerical differentiation based on finite differencing is prone to 
round-off errors, while symbolic differentiation is in theory exact but requires 
the entire computer source code to be compiled to a mathematical expression. 
In contrast, automatic differentiation allows exact computation of gradients 
(up to the standard floating point error) without the need to compile source code
to a symbolic mathematical expression. 
One of the ways to implement AD this is by replacing calculations on the real numbers with corresponding 
``dual numbers'' in which the second coordinate eventually stores the derivative output. 
We refer readers to \cite{baydin2018autodiff} for more information on automatic differentiation 
and a more comprehensive review.

When using automatic differentiation, it is important to understand that it comes 
in two main differentiation modes: forward and reverse mode.  
In short, the modes differ in the order in which the chain rule is applied: forward 
mode works toward the outside from the inside, while reverse mode works in the opposite 
direction. This has important computational consequences (with respect to both time and memory) 
when differentiating a function $f:\reals^n \to \reals^m$, where $n \ll m$ or $m \ll n$. 
In general, if $n \ll m$, it is preferred to perform forward-mode AD, whereas 
reverse-mode AD is generally preferred when $m \ll n$ 
(at the cost of a higher memory consumption in the latter case).

Automatic differentiation is available in most programming languages as an additional package 
or extension. 
For instance, Python and NumPy code can be automatically differentiated with JAX.
In Julia, several packages are available for reverse- or forward-mode differentiation, 
such as \texttt{Zygote}, \texttt{Enzyme}, and \texttt{ForwardDiff}.
In C++, the \texttt{autodiff} library allows users to perform automatic differentiation.

\subsection{Mini-batching}
\label{sec:mini-batch}

Consider the special but important case where the objective function $f(x, \phi)$ can be
written as a sum over $N$ terms, so that
\[ \label{eq:sum} f(x, \phi) = \sum_{i=1}^N f_i(x, \phi). \]
Typically, $N$ is the number of observations, assumed to be large. 
When \cref{eq:sum} holds, 
the gradient can be approximated by sampling an index $I$ 
uniformly at random in the set $\{1, 2, \dots, N\}$ and by using only the $I$-th 
term with weighting $N$, resulting in the estimator
\[ \hat G = N \hat G_I, \]
where $\hat G_i$ is the gradient estimator for term $i$. 
This construction is motivated by the following: if each $\hat G_i$ is 
unbiased for $-\nabla f_i(x, \phi)$, then $\hat G$ is also unbiased. That is,
\[ 
  \ex[\hat G] 
  = N \cdot \ex [ \ex[\hat G_I | I] ] 
  = -N \sum_{i=1}^N \frac{1}{N} \nabla f_i(x, \phi) 
  = -\nabla f(x, \phi). 
\]
This can be generalized to picking more than one term at a time; 
each group of such terms is called a \emph{mini-batch}. 
This yields a trade-off: smaller mini-batches are faster to evaluate but are more noisy, 
while larger mini-batches can be more expensive to evaluate but are generally less noisy. 
When selecting a mini-batch size, the given hardware constraints are a key consideration. 
In particular, when computation is done in a vectorized fashion, i.e., 
by doing all the items in the mini-batch in parallel, there is a  
maximum number of items that can be processed in parallel. 
For example, when vectorization is done on a GPU, the constraints will 
be the number of cores in the GPUs as well as its available memory.  

The idea of mini-batching illustrates an important general principle: it is  
possible to trade-off more computational cost per stochastic gradient evaluation 
for a reduced variance. For any gradient estimator, one can simulate it $B$ times, 
$\hat G^{(1)}, \hat G^{(2)}, \dots, G^{(B)}$, and use the average, 
$\sum_b  \hat G^{(b)}/B$, 
to create a  
lower variance but computationally more expensive gradient estimator.

\subsection{Stochastic gradient descent algorithms}

We now turn to the question of the choice of optimization algorithm 
for approaching MOI problems given by \cref{eq:key,eq:min}. 

We take as a starting point the stochastic gradient algorithm. 
When using an optimal step size selection, this algorithm converges at the rate of 
\[  \ex[ f(\phi_t) ] - \min_\phi f(\phi)   = O(1/\sqrt{t}), \] 
for convex $f(\phi)$ and Lipschitz continuous $g(\phi)$ \cite{nemirovski_robust_2009}. 
In fact, the above rate is matched with well-known lower bound results, established 
for \iid gradient oracles 
\cite{agarwal_information-theoretic_2012,nemirovskii_problem_1983}. In other words,  
there exists MOI problems where SGD is provably optimal in terms of the 
convergence rate in $t$.
Of course, for other more specific problem classes, specialized algorithms can outperform 
SGD in theory and practice. 
For example, such specialized algorithms have been developed in the 
case where the stochasticity of the gradient estimator only comes from the mini-batch selection 
(i.e., where $\hat g_i = -\nabla f_i$ in the notation of \cref{sec:mini-batch} in \cref{eq:sum})  
\cite{schmidt_minimizing_2017}, a setup that however excludes most non-trivial MOI problems 
discussed in this chapter.
Moreover, since the mini-batch selection setup 
is central to many deep learning methods, 
the stochastic optimization literature currently puts a strong emphasis on that special case. 
As the MOI setup has distinct characteristics, it is important to be aware that 
popular recommendations in that literature (both from theory and practical experience) 
may not necessarily be optimal in 
certain MOI problems.

This initial discussion suggests two follow-up questions. 
First, before recommending SGD as a generic, default choice for MOI, one
has to discuss the selection of the learning rate $\gamma_t$ (also known as the step size), 
as SGD's theoretical properties and practical effectiveness heavily depend on it. 
Second, what algorithms beyond SGD should be considered to attack specific classes 
of MOI problems?

\subsubsection{Selection of SGD learning rates}

As described in more detail in \cref{sec:theory}, there is a tension when setting $\gamma_t$, 
where values that are too large lead to instability while values that are too small lead to slow progress. 
Classical treatments of SGD consider parametric forms of decay for  $\gamma_t$, 
for example of the form $\gamma_0 (1 + C \gamma_0 t)^{-\alpha}$, 
with $\alpha \in (1/2, 1]$, see e.g.,  \cite[Section 18.5.2]{montavon_neural_2012}, and 
values for $\alpha$ closer to the lower bound of the interval often 
preferred, see e.g., \cite[Section 4.2.2]{andrieu2008tutorial}.
Another special case of parametric ``decay'' is to use a constant step size, 
e.g., \cite[Section 2.2]{nemirovski_robust_2009}.

In many MOI applications the requirement is not to optimize to high accuracy 
but rather to quickly get in a neighborhood of the solution (for example, due to model 
mis-specification; or when the learned parameters are only used to assist MCMC sampling). 
SGD is a good fit for this kind of situation as its fast initial speed of convergence is often 
considered its strength, however it can only achieve this when the learning rate is good 
in a non-asymptotic sense. 
This motivates the development of automatic step size algorithms, which sidestep 
the need to fix a parametric form for $\gamma_t$. 
The main approaches explored in that large literature can be roughly categorized 
in the following types: methods that collect summary statistics on the points and gradients observed 
so far, methods that perform gradient descent on the step size, and finally, methods 
that use a line search algorithm---see e.g., \cite[Section 1.1]{carmon_making_2022} for 
a recent literature review. 
Most previous approaches, however, either void SGD's theoretical guarantees---for 
example, subsequent analysis \cite{j.2018on} of the popular Adam optimizer 
\cite{kingma_adam_2015} identified an 
error in the proof of convergence and provided an example of non-convergence 
on a convex function---or focus on setups that may not be representative of certain MOI problems,
e.g.,  where stochasticity comes from mini-batch selection 
or using specific characteristics of deep learning training. 
Another recent approach that shows promise both empirically and theoretically 
(in the \iid setting) 
is the Distance over Gradients (DoG) method \cite{ivgi_dog_2023}. However, similarly 
to most step size adaptation methods, DoG has not be analyzed in the Markovian setting 
yet. 

We note that progressively increasing the size of the mini-batch can be used as an 
alternative (or in combination) to a decrease in step size, a technique sometimes 
known as ``batching'' \cite{friedlander_hybrid_2012}, and is related to the 
round-based scheme described in \cref{sec:example}. Batching can be generalized to 
methods that progressively decrease the variance of $\hat G_t$, and in that optic 
relates to Sample Average Approximation (SAA) methods \cite{kim_guide_2015}. 
In some cases, this variance may also decrease naturally as a consequence of 
the form of specific problems, see e.g., \cite{chen2023coreset}. 

It should also be noted that often these methods do not completely 
remove the need for tuning SGD's learning rate, but 
instead reduce it to a small number of parameters, such as an initial step size. 
One can then use a grid over the remaining parameter(s)  and compare the 
objective function values obtained.

\subsubsection{Other algorithms}

The vanilla SGD algorithm can be improved in many ways and forms the basis 
for several more advanced optimization algorithms. 
A first common modification is to use a different learning rate for each dimension, 
to take into account that they might require different scales.  
A flurry of methods have been developed for doing this, including 
Adagrad \cite{duchi2011adaptive}, AMSGrad \cite{reddi2019convergence}, and
Adam \cite{kingma_adam_2015}.
A building block for this line of work is to incorporate a ``momentum'' to the optimizer, 
which after unrolling the updates leads to 
\[ \phi_{t+1} &= \phi_t + \gamma g(X_t, \phi_t) + \beta (\phi_t - \phi_{t-1}), \]
a method known as momentum gradient descent or the heavy ball algorithm \cite{polyak64}. 
See also \cite{nesterov_method_1983} and \cite[Section 2]{sutskever_importance_2013} 
for the closely related notion of accelerated gradient.

Another well-known perspective to using a different learning rate for each dimension is 
that it acts as a diagonal pre-conditioning matrix, see, e.g., \cite[Section 5]{bordes_sgd-qn_2009}. 
The natural generalization is to look at full pre-conditioning, leading to so-called second order methods. 
When the pre-conditioning matrix is based on the Hessian matrix this leads to Newton's method.  
However, the general wisdom 
is that the high cost per iteration involved with manipulating the Hessian matrix does not justify 
the faster convergence when the dimensionality is large. 
In fact, the trade-off is even more against second-order methods in the stochastic optimization 
setting compared to the deterministic one due to the fact that noise makes the 
quadratic Hessian approximation often less reliable. 
An important exception to this rule arise when it is possible to implicitly 
precondition by careful selection of the variational family parameterization, hence 
avoiding the costly manipulation of a pre-conditioning matrix. 
We give an example in \cref{sec:moment-as-precond}.

Another attractive variant of pre-conditioning is to approximate the Hessian matrix rather than 
computing it exactly, leading to quasi-Newton or more broadly, ``1.5 order'' methods. 
We refer readers to \cite{guo_overview_2023} for a recent review of these methods 
in the context of stochastic optimization. 
The most common way to form the Hessian approximation is to use a low-rank matrix updated 
from the history of previously computed gradients, leading to the L-BFGS algorithm \cite{liu_limited_1989}.
Recently, alternatives based on early stopping of the conjugate gradient algorithm used in an inner loop 
\cite{dembo_truncated-newton_1983,nash_survey_2000}
have regained popularity in the stochastic context, see e.g., \cite{schulman_trust_2015}. 
Part of the reason for this resurgence is the development of sophisticated 
libraries for automatic differentiation (\cref{sec:autodiff})
that can efficiently compute Hessian vector products using a combination of 
reverse-mode and forward-mode automatic differentiation \cite{jax2018github}.

Polyak averaging is a technique for stabilizing stochastic gradient 
descent methods \cite{ruppert_efficient_1988,polyak_new_1990}. 
The idea is to run standard SGD, and to return the average of the 
parameters instead of the last parameter visited. 
This can also be modified to incorporate a ``burn-in,'' 
see e.g., \cite{durmus_finite-time_2023}.  
This technique is useful in particular because methods based on stochastic gradient 
descent often converge to within a ball, at which point the iterates move randomly 
within the ball due to noise from the gradient estimates. The Polyak averaging 
technique can partially cancel this noise.

\subsection{Stabilization of gradient descent}
\label{sec:gradient_stabilization}

Estimates of the gradient $\nabla f$ may have a large variance even if they are unbiased. 
We introduce two simple techniques that are useful for stabilizing
gradient descent: control variates and Rao-Blackwellization.
The method of control variates is a technique that can be used to reduce the variance 
of estimates of an unkown quantity of interest. 
Let $g = -\nabla f$ be the gradient of interest and $\hat g = -\hat\nabla f$ an unbiased
estimate of $g$. That is, $\ex[\hat g] = g$. By introducing a \emph{control variate}, 
$\hat c$, which is a random variable with a \emph{known} expected value $c$, one 
can substantially reduce the variance of an estimate of $g$ provided that 
$\abs{\cor[\hat g, \hat c]}$ is high. The form of an estimate of $g$ given by the method 
of control variates is given by $\hat g + k (\hat c - c)$, where $k$ is a tunable 
constant. To see that such an estimator can reduce the variance of $\hat g$, observe that 
\[
  \var[\hat g + k (\hat c - c)] 
  = \var[\hat g] + k^2 \var[\hat c] + 2k \cov[\hat g, \hat c].
\]
Solving for $k$ to minimize the magnitude of this expression, we find that 
\[
  k^\star &= -\cov[\hat g, \hat c]/\var[\hat c] \\
  \var[\hat g + k^\star (\hat c - c)] &= (1 - \cor[\hat g, \hat c]^2) \var[\hat g].
\]
In general, $k^\star$ is not known beforehand, but can sometimes be estimated using 
Monte Carlo samples $\hat{g}_1, \ldots, \hat{g}_T$ and $\hat{c}_1, \ldots, \hat{c}_T$. 
Similarly, even if the value of $c$ is unknown, this quantity can also be 
estimated using Monte Carlo samples.

In certain circumstances, Rao-Blackwellization techniques
\cite{casella1996rao,ranganath2014black} offer substantial variance
reduction without any tuning parameters.
Let $X = (X_1, X_2) \sim \pi_\phi$ so that $g(X,\phi)$ is an unbiased estimate of $\nabla_\phi f(\phi)$.
By the Rao-Blackwell theorem, the conditional expectation
\[
  \ex_{\pi_\phi(X_2 \mid X_1)} \left[ g((X_1,X_2), \phi) \right]
\]
is a lower variance estimate of the gradient. 
This estimator is useful provided that it does not depend on the unknown quantity 
$\nabla_\phi f(\phi)$ (i.e., we condition on a sufficient statistic), 
and that the conditional expectation of $g$ can be computed analytically.
Such cases occur frequently in (reverse KL) variational inference settings
when the variational distribution over $X_1, X_2$ is the product of independent
densities \cite{ranganath2014black}.

\section{Theoretical analysis}
\label{sec:theory} 

In this section we provide a brief introduction to the analysis of the stability and convergence
behaviour of MOI algorithms. 
Having unified various problems at the intersection of MCMC and ML, such theory 
can be applied to many different settings.
When using the adaptive algorithm \cref{eq:rm} to solve an MOI problem \cref{eq:key},
there are a few questions a user may be interested in, depending on the particular case under consideration:
\bitems
\setlength{\itemsep}{0pt}
\item Do the parameters $\phi_t$ remain in a compact region of $\reals^m$?
\item Do the parameters $\phi_t$ converge to a solution $\phi^\star \in \reals^m$ of \cref{eq:key} (or \cref{eq:min})?
\item If the parameters $\phi_t$ converge to some $\phi^\star\in\reals^m$, at what rate do they converge?
\item Are the states $X_t$ useful for estimating expectations under some distribution?
\eitems

Answers to the above questions are remarkably technically challenging to obtain at a reasonable level of generality,
and failure modes abound even with restrictive assumptions. In order to focus the discussion of this section on
the core interesting challenges that arise from MOI problems---as opposed to technicalities---we will assume
that $\Phi = \reals^m$, that $g(\phi) = -\grad f(\phi)$ for $g,f$ in \cref{eq:key,eq:min}, that $f$ is twice differentiable,
and that each kernel $\kappa_\phi$ has a single stationary distribution $\pi_\phi$. A much more thorough, general,
and rigorous (but less accessible) treatment of this material can be found in the literature on 
stochastic approximation; see \cite{andrieu_stability_2005,andrieu_ergodicity_2006,andrieu_expanding_projections_2014}
and especially \cite{andrieu_stability_2005} for a good entry point into that literature.

\subsection{Counterexamples, assumptions, and results}

Failures of the adaptive algorithm occur due to one of a few causes: poor properties of the objective function $f$,
insufficiently controlled noise in the random estimate $g(X, \phi) \approx g(\phi)$ given $X\dist \pi_\phi$, 
a poorly chosen step size sequence $\gamma_t$,
and bias of the estimate $g(X, \phi)$ due to slow mixing of the kernel $\kappa_\phi$.
Note that only the last of these sources of failure involve the Markovian nature of the problem; the first three
also occur in the classical Robbins--Monro \cite{robbins_monro_1951} setting with independent draws $X \dist \pi_\phi$. 

We split our analysis into three nested parts and gradually introduce assumptions required 
for each setting. We first consider a setup where the gradients 
are deterministic, corresponding to a review of standard gradient descent theory. 
Then, we consider the case of unbiased gradient estimates and introduce appropriate 
assumptions that guarantee convergence in the presence of gradient noise; these 
results are essentially a review of the theory underlying stochastic gradient descent. 
Finally, we assume that the noise in gradient estimates is Markovian. This final 
setting requires the most assumptions in order to guarantee convergence to an optimum. 

\begin{figure}
\begin{subfigure}{0.49\textwidth}
\centering\includegraphics[width=\linewidth]{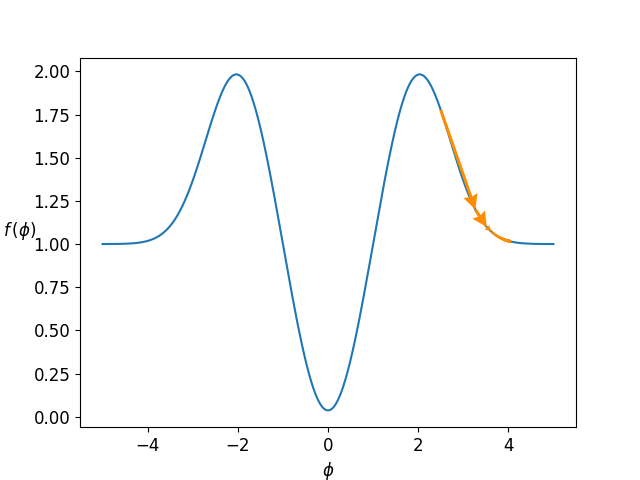}
\caption{}\label{fig:wiggle}
\end{subfigure}
\begin{subfigure}{0.49\textwidth}
\centering\includegraphics[width=\linewidth]{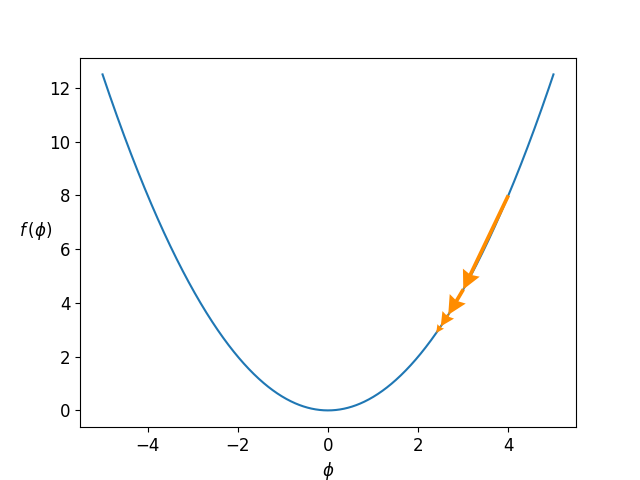}
\caption{}\label{fig:stuck_quadratic}
\end{subfigure}
\begin{subfigure}{0.49\textwidth}
\centering\includegraphics[width=\linewidth]{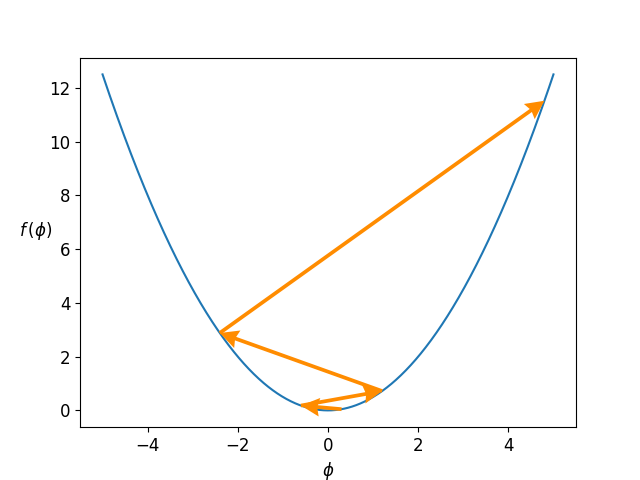}
\caption{}\label{fig:diverge_quadratic}
\end{subfigure}
\begin{subfigure}{0.49\textwidth}
\centering\includegraphics[width=\linewidth]{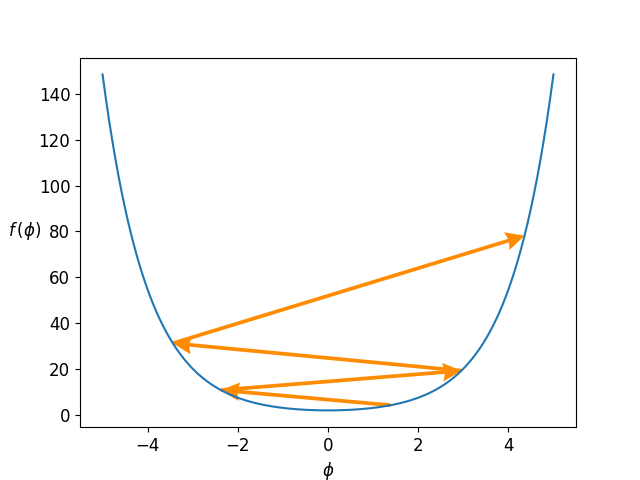}
\caption{}\label{fig:cosh}
\end{subfigure}
\caption{Examples of failures of the adaptive algorithm \cref{eq:rm} in a deterministic setting.
Orange arrow segments depict the sequence of iterates.
\cref{fig:wiggle}: $f$ is given by \cref{eq:wiggle}; the adaptive algorithm is led away from a solution by a decaying right tail.
\cref{fig:stuck_quadratic}: $f$ is given by \cref{eq:quadratic}; the step size sequence decays too quickly, and the adaptive algorithm gets stuck.
\cref{fig:diverge_quadratic}: $f$ is given by \cref{eq:quadratic}; the step size sequence is too aggressive, and the adaptive algorithm becomes unstable.
\cref{fig:cosh}: $f$ is given by \cref{eq:cosh}; the step size sequence is reasonable, but the function $f$ is not Lipschitz smooth, leading again to instability.}
\end{figure}

\subsubsection{Deterministic setup}

We begin with the simplest case where the kernel $\kappa_\phi$ mixes quickly
and the estimates $g(X, \phi)$ have little noise. Mathematically, we model this
extreme case by considering a deterministic setup assuming there is no
estimation noise at all (i.e., for all $\phi \in \Phi$, $x \in \fcX$,  $g(x,
\phi) = g(\phi)$) in the update \cref{eq:rm}. Even in this case, it is possible
that the adaptive algorithm produces a sequence $\phi_t$ that not only fails to
converge to a solution of \cref{eq:key}, but diverges entirely; there is little
hope for a stochastic version of the adaptive algorithm unless we first address
these counterexamples. For example, consider the function $f:\reals\to\reals$ shown in
\cref{fig:wiggle},
\[
	f(\phi) = 1-e^{-\phi^2} + e^{-(\phi-2)^2} + e^{-(\phi+2)^2}, \label{eq:wiggle}
\]
which has a single local minimum at $\phi^\star = 0$.
But if \cref{eq:rm} is initialized at $\phi_0 = 2.5$, for any reasonable choice of step sequence $\gamma_t$, we have $\phi_t \to \infty$.
To prevent this failure, we need to assume that $g$ is such that the noise-free dynamics stabilizes around solutions
$\Phi^\star = \{\phi : g(\phi) = 0\}$ of \cref{eq:key}. In what follows, let $d(x, A)$ for $x\in\reals$, $A\subseteq \reals$ be defined
as $d(x, A) = \inf_{y\in A} \|x-y\|$.
\bassump \label{assump:stableg}
$\inf_\phi f(\phi) = \underline f \in \reals $, $\Phi^\star \neq \emptyset$, and if $\cbra{\phi_t}_{t \geq 0}$ is 
a sequence such that $g(\phi_t) \to 0$, then $d(\phi_t, \Phi^\star) \to 0$.
\eassump
\noindent It is also important to carefully select the step size sequence $\gamma_t$ to avoid instabilities 
and converging to a point not equal to any solution in $\Phi^\star$. For example, consider the quadratic objective function 
$f:\reals\to\reals$ shown in \cref{fig:stuck_quadratic,fig:diverge_quadratic},
\[
f(\phi) = \frac{1}{2}\phi^2. \label{eq:quadratic}
\]
If we initialize at $\phi_0 \neq 0$ and use the quickly-decaying step size sequence $\gamma_t = 1/(t+1)^2$, we find that
as $t\to\infty$,
\[
\phi_t = \phi_0 \prod_{k=0}^{t-1} (1-\gamma_k) \approx \phi_0 e^{-\sum_{k=0}^{t-1} \gamma_k} = \phi_0 e^{-\sum_{k=1}^t\frac{1}{k^2}} \to \phi_0 e^{-\pi^2/6} \neq 0, \label{eq:phitstuck}
\]
and hence the parameter sequence never approaches $\Phi^\star$ because the initialization bias does not decay to 0 (see \cref{fig:stuck_quadratic}).
Examining \cref{eq:phitstuck}, to avoid this behaviour, we need that the step sizes are aggressive enough such that their sum diverges. 
\bassump \label{assump:stepsumdiverge}
The step size sequence satisfies
\[
\sum_{k=0}^t \gamma_k \to \infty \quad\text{as}\quad t\to\infty.
\]
\eassump
\noindent On the other hand, we need the step sizes $\gamma_t$ to be small enough to avoid instabilities in $\phi_t$.
For example, consider the same objective function
except with fixed step size $\gamma_t = 3$. Then $\phi_{t+1} = -2\phi_t$, and so for any initialization $\phi_0 \neq 0$ (see \cref{fig:diverge_quadratic}),
\[
|\phi_t| = |\phi_0 (-2)^t| \to \infty \quad \text{as} \quad t\to\infty.
\]
Instabilities of this kind can occur even with reasonably decaying step sizes if $g(\phi)$ grows
too quickly. For example, consider the function $f:\reals\to\reals$ given by
\[
f(\phi) = e^{\phi} + e^{-\phi}. \label{eq:cosh}
\]
Initializing the algorithm at $\phi_0 = 1.39$ and using step sizes $\gamma_t = 1/(t+1)$
causes the sequence of parameters $\phi_t$ to diverge (see \cref{fig:cosh}). To prevent such instabilities,
we require that the step sizes are set in such a manner that they are compatible with the (Lipschitz) smoothness of the objective function;
generally speaking, a smoother objective enables more aggressive steps. 
\bassump \label{assump:smooth}
There exists an $L < \infty$ such that 
\[
\sup_{\phi\neq \phi'} \frac{\Vert g(\phi) - g(\phi')\Vert}{\Vert\phi-\phi'\Vert} \leq L, \quad\text{and}\quad \sup_t \gamma_t < \frac{1}{L}.
\] 
\eassump

\noindent It turns out these are the only pathologies in the deterministic case; \cref{thm:deterministic} shows
that as long as they are all circumvented using the aforementioned assumptions, the adaptive algorithm 
given by \cref{eq:rm}---which, in the deterministic case is just vanilla gradient descent---converges to the 
set of solutions $\Phi^\star$ of \cref{eq:key}.
The proof is inspired by that of \cite[Theorem 2]{lei_nonconvex_sgd_2020} and \cite[Theorem 2.1]{ghadimi_nonconvex_sgd_2013},
but focuses on the deterministic setting.
\bthm\label{thm:deterministic}
Suppose \cref{assump:stableg,assump:stepsumdiverge,assump:smooth} hold, and 
for all $x\in\fcX$, $\phi\in\Phi$, $g(x,\phi) = g(\phi)$. Then,
\[
g(\phi_t) \to 0 \quad \text{and} \quad d(\phi_t, \Phi^\star) \to 0 \quad \text{as} \quad t\to\infty,
\]
and
\[
\min_{k\in\{0, \dots, t\}} \|g(\phi_t)\|^2 \leq \frac{f(\phi_0)-\underline f}{\sum_{k=0}^t \gamma_k\lt(1-\frac{\gamma_k L}{2}\rt)}.
\]
\ethm
\bprf
By Taylor's theorem, for some $\xi_t \in [\phi_t, \phi_{t+1}]$, recalling that $g(\phi) = -\nabla f(\phi)$
and $\phi_{t+1} = \phi_t + \gamma_t g(\phi_t)$, we have
\[
f(\phi_{t+1}) 
&= f(\phi_t) - g(\phi_t) ^T(\phi_{t+1} - \phi_t) + \frac{1}{2}(\phi_{t+1}-\phi_t)^T \grad^2 f(\xi_t)(\phi_{t+1}-\phi_t)\\
&\leq f(\phi_t) - \gamma_t \|g(\phi_t)\|^2 + \frac{\gamma_t^2L}{2}\|g(\phi_t)\|^2. \label{eq:detthmrec1}
\]
So
\[
\gamma_t \lt(1-\frac{\gamma_t L}{2}\rt)\|g(\phi_t)\|^2 &\leq f(\phi_t)-f(\phi_{t+1})\\
\sum_{k=0}^t \gamma_k \lt(1-\frac{\gamma_k L}{2}\rt)\|g(\phi_k)\|^2 &\leq \sum_{k=0}^t f(\phi_k)-f(\phi_{k+1}) = f(\phi_0) - f(\phi_{t+1}) \leq f(\phi_0) - \underline f.
\]
Therefore,
\[
\min_{k\in\{0,\dots,t\}}\|g(\phi_k)\|^2 
&\leq \frac{f(\phi_0)-\underline f}{\sum_{k=0}^t \gamma_k \lt(1-\frac{\gamma_k L}{2}\rt)}.
\]
By modifying the previous telescoping sum technique to start at $k > 0$ instead of at $k=0$, we 
see that $\liminf_{t\to\infty} \|g(\phi_t)\|^2 = 0$.
Now suppose $\limsup_{t\to\infty} \|g(\phi_t)\|^2 > 0$.
Therefore, there exists an $\epsilon > 0$ and infinitely
many segments of indices $t_n, t_n+1,\dots, T_n$, $T_n > t_n$, such that
\[
\|g(\phi_{t_n})\| &\geq 2\epsilon, &
\|g(\phi_{T_n})\| &\leq \epsilon, & \text{and}\quad 
\forall t_n < t < T_n, \quad \|g(\phi_t)\| &\geq \epsilon.
\]
Therefore,
\[
\epsilon^2 &\leq \epsilon\lt(\|g(\phi_{t_n})\| -  \|g(\phi_{T_n})\|\rt)\\
&\leq \epsilon\|g(\phi_{t_n})- g(\phi_{T_n})\|\\
&\leq \epsilon \sum_{k=t_n}^{T_n-1} \|g(\phi_{k+1})- g(\phi_{k})\|\\
&\leq L \epsilon \sum_{k=t_n}^{T_n-1} \|\phi_{k+1}- \phi_{k}\|\\
&\leq L \epsilon\sum_{k=t_n}^{T_n-1} \gamma_k \|g(\phi_k)\|\\
&\leq L \sum_{k=t_n}^{T_n-1} \gamma_k \|g(\phi_k)\|^2.
\]
Since $\frac{1}{2} \leq \lt(1- \frac{\gamma_kL}{2}\rt)$,
\[
\frac{\epsilon^2}{2L} \leq \sum_{k=t_n}^{T_n-1} \gamma_k \lt(1- \frac{\gamma_kL}{2}\rt)\|g(\phi_k)\|^2.
\]
Using this result in the previous telescoping sum proof technique yields
\[
\frac{\epsilon^2}{2L} \leq f(\phi_{t_n}) - f(\phi_{T_n}).
\]
The above suggests that $f(\phi_t)$ undergoes infinitely many
decreases of size at least $\epsilon^2/2L$, and hence is unbounded below, 
which is a contradiction. Therefore, we have that $\|g(\phi_t)\| \to 0$, and thus $g(\phi_t) \to 0$ and $d(\phi_t, \Phi^\star) \to 0$.
\eprf

\subsubsection{Independent noise}

Next, we identify failure modes caused by the use of noisy estimates $g(X, \phi)$ of $g(\phi)$ in the adaptive algorithm \cref{eq:rm}.
Prior to moving onto a fully general setting, we consider again the simpler case where the kernel $\kappa_\phi$ always mixes quickly; we
model this setting mathematically by assuming $\kappa_\phi$ corresponds to independent draws from $\pi_\phi$. 
The first failure mode relates again to poor choices of the step size sequence.
In particular, consider the smooth, strongly convex function $f:\reals\to\reals$ with unique minimum $\phi^\star = 0$, 
\[
	f(\phi) = \frac{1}{2}\phi^2, \label{eq:quadratic_again}
\]
with estimate given by $g(X, \phi)$, $X\dist \pi_\phi= \distNorm(0,1)$, where
\[
	g(x, \phi) &= -\phi + x.
\]
In this case we can write the sequence $\phi_t$ in closed form:
\[
	\phi_t &= \phi_0 \prod_{k=0}^{t-1} (1-\gamma_k) + \sum_{k=0}^{t-1}\gamma_k X_k \prod_{j=k+1}^{t-1}(1-\gamma_j).
\]
\begin{figure}
\centering\includegraphics[width=0.6\textwidth]{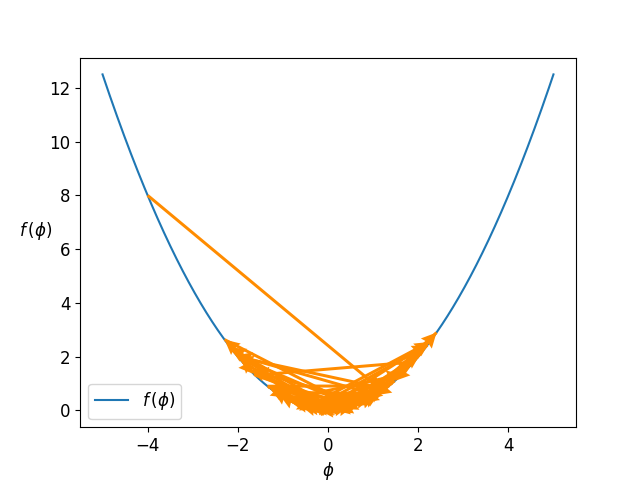}
\caption{An example of a failure of the adaptive algorithm \cref{eq:rm} due to noise. 
Orange arrow segments depict the sequence of iterates.
The objective
$f$ is given by \cref{eq:quadratic_again}. Because the step size sequence is a constant, the perturbations
due to noise do not decay adequately and the sequence does not converge.} \label{fig:noiseball}
\end{figure}

We can satisfy the earlier \cref{assump:smooth} by setting $\gamma_t = 1/2$; the initialization bias converges
geometrically to 0 as $t\to\infty$. However, with a constant step size, the noisy estimates perturb the sequence $\phi_t$ enough such
that it never converges to $\phi^\star$, but rather bounces around within a ball containing $\phi^\star$ (see \cref{fig:noiseball}). 
Indeed, $\phi_t$ is normally distributed with mean and second moment given by
\[
\ex[\phi_t] &= \phi_0\prod_{k=0}^{t-1} (1-\gamma_k) = \phi_0 2^{-t}\\
\ex[\phi_t^2] &= \phi^2_0\prod_{k=0}^{t-1} (1-\gamma_k)^2 + \sum_{k=0}^{t-1}\gamma^2_k \prod_{j=k+1}^{t-1}(1-\gamma_j)^2
= \phi_0^2 4^{-t} +\frac{1-4^{-t}}{3},
\]
and hence $\phi_t$ does not converge to a solution, but instead $\phi_t \convd \distNorm(0, 1/3)$ as $t\to\infty$. 
A more severe issue can occur if the variance of the noise is unbounded. There are many ways to construct examples
in this situation that result in failure of the adaptive algorithm. For example, consider the same problem
except that $\pi_\phi = \distNorm(0, e^{2\phi})$. Then, for $Z_t \distiid \distNorm(0,1)$,
\[
\phi_{t+1} = (1-\gamma_t)\phi_t + \gamma_t e^{\phi_t} Z_t.
\]
Therefore, 
\[
\pr\lt(|\phi_{t+1}| > 2|\phi_t| \given \phi_t\rt) 
&= 2 - F\lt(|\phi_t|e^{-\phi_t}\frac{3-\gamma_t}{\gamma_t}\rt) 
  - F\lt(|\phi_t|e^{-\phi_t}\frac{1+\gamma_t}{\gamma_t}\rt),
\]
where $F(x)$ is the CDF of the standard normal.
Suppose we initialize $\phi_0 = 2$ and use $\gamma_t = 1/(2(t+1))$. 
Then, if $|\phi_{k+1}| > 2|\phi_k|$ for all $k=0, \dots, t-1$,
we have that $|\phi_t| > 2^{t+1}$. Therefore
\[
\pr\lt(|\phi_{t+1}| > 2|\phi_t| \given \min_{k < t} \frac{|\phi_{k+1}|}{|\phi_k|} > 2\rt)
&\geq 2 - F\lt(O(t 2^t e^{-2^t})\rt) - F\lt(O(t 2^t e^{-2^t})\rt)\\
&= 1- O(t 2^t e^{-2^t}),
\]
where the last line follows for large $t$ from Lipschitz continuity of $F$ at $0$.
Therefore,
\[
\pr\lt(\forall t, |\phi_t| > 2^{t+1}\rt) > 0,
\]
and neither stability nor convergence are guaranteed.
To address this, we essentially need to enforce that the variance of $\gamma_t g(X_t, \phi_t)$ decays faster
than the mean as $t\to\infty$. Although there are many ways to accomplish this, here we bound the second
moment with a constant and a term that may grow as $g$ does.
\bassump\label{assump:boundednoise}
There exist $a, b > 0$ such that 
\[
\ex\lt[\|g(X_t, \phi_t)\|^2\rt] \leq a + b\|g(\phi_t)\|^2, \quad \sup_t \gamma_t < \frac{1}{Lb}, \quad \sum_{t=0}^\infty \gamma_t^2 < \infty.
\]
\eassump
This is the only additional assumption required to obtain convergence in \cref{thm:independentnoise},
which provides very similar guarantees to those in \cref{thm:deterministic}, but in the stochastic
setting with independent noise $X_t \dist \pi_\phi$. 
\bthm\label{thm:independentnoise}
Suppose \cref{assump:stableg,assump:stepsumdiverge,assump:smooth,assump:boundednoise} hold, 
and for all $x'\in\fcX$, $\kappa_\phi(\dee x, x') = \pi_\phi(\dee x)$.
Then,
\[
g(\phi_t) \convp 0, \qquad d(\phi_t, \Phi^\star) \convp 0,  \qquad t\to\infty,
\]
and
\[
\min_{k\in\{0,\dots,t\}}\ex \|g(\phi_k)\|^2 &\leq \frac{f(\phi_0) - \underline f +  \frac{\sum_{k=0}^t\gamma_k^2 a L}{2}}{\sum_{k=0}^t \gamma_k \lt(1-\frac{\gamma_k b L}{2}\rt)}.
\]
\ethm

\bprf
The proof is nearly identical to that of \cref{thm:deterministic}, 
except that we use \cref{assump:boundednoise} to bound $\ex \|g(X_t, \phi_t)\|^2$ above with $a + b \|g(\phi_t)\|^2$.
We also note that $\sum_{k=t_n}^{T_n-1} \gamma_k^2 a L/2$ converges to 0 as $n \to \infty$. 
Finally, we show that $g(\phi_t) \convp 0$ implies that $d(\phi_t, \Phi^\star) \convp 0$
by using the property that any subsequence has a further subsequence where $g\to 0$ almost surely 
and hence $d\to 0$ almost surely on that same subsequence, yielding the desired convergence 
in probability on the full sequence.
\eprf

\subsubsection{Markovian noise}

We finally consider failures of the adaptive algorithm \cref{eq:rm} caused by poor behaviour of the Markov kernel $\kappa_\phi$.
At a high level, as long as the mixing of $\kappa_\phi$ is uniformly well-behaved over the domain $\Phi$,
the draws from $\kappa_\phi$ will look sufficiently like independent draws to obtain results similar to the independent noise setting.
Issues can occur, however, when the $\kappa_\phi$ are not uniformly well-behaved; this is true even when the objective $f$ and distribution $\pi_\phi$ are ideal,
and the $\kappa_\phi$ are pointwise well-behaved.
For example, consider
\[
	f(\phi) = \ex\lt[f(X, \phi)\rt], \quad X\dist \distUnif\{+, -\}, \label{eq:strcvx}
\]
where
\[
	f(+, \phi) &= \ind[\phi \leq 0] (\phi-1)^2 + \ind[\phi > 0](1-2\phi)\\
	f(-, \phi) &= \ind[\phi \leq 0] (1+2\phi) + \ind[\phi > 0] (1+\phi)^2.
\]
\begin{figure}
\centering\includegraphics[width=0.6\textwidth]{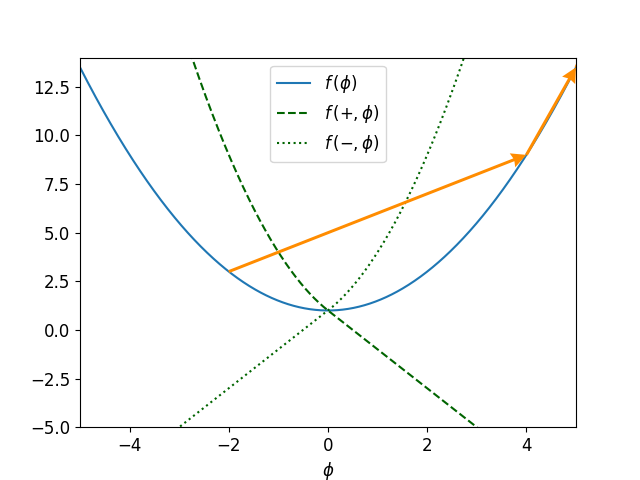}
\caption{An example of a failure of the adaptive algorithm \cref{eq:rm} due to non-uniform kernel
mixing behaviour. Orange arrow segments depict the sequence of iterates. The objective $f$ is given by \cref{eq:strcvx}. The algorithm
is initialized in the $+$ state (green dashed line),
and so the iterates follow the negative gradient and proceed in the positive direction. However, the kernel mixes increasingly more slowly
for parameters of larger magnitude;
eventually the kernel becomes stuck in the $+$ state and the iterates diverge to $+\infty$.} \label{fig:strcvx}
\end{figure}
As shown in \cref{fig:strcvx}, the objective $f$ is strongly convex and Lipschitz smooth and has a unique minimum at $\phi^\star =0$;
the two components $f(\cdot,\phi)$ are likewise convex and Lipschitz smooth.
Inspired by an example from Younes \cite[Sec.~6.3]{younes_decreasing_ergodicity_1999}, we use the kernel
$\kappa_\phi$ that toggles the state with probability $\exp(-\exp(|\phi|))$ and remains in the current
state otherwise. For all $\phi\in\reals$, the invariant distribution of this kernel is $\distUnif\{+, -\}$ 
and each kernel $\kappa_\phi$ is uniformly geometrically ergodic (with respect to the 
initialization $x \in \cbra{+, -}$), as desired. However, the ergodicity is not uniform over $\phi$.
In particular, if we initialize the system at $\phi_0 = 0$, $X_0 = +$, and use the step size $\gamma_t = 1/(t+1)$,
we have that
\[
\pr\lt(\forall t, X_t = +\rt) &= \prod_{t=0}^\infty \lt(1-e^{-e^{2\sum_{k=0}^t \gamma_k}}\rt) 
\approx
\prod_{t=1}^\infty \lt(1-e^{-t^2}\rt) > 0.
\]
Therefore, the probability that the state never switches---and hence $\phi_t \to \infty$---is positive, and neither
stability nor convergence are guaranteed. The remedy to this problem, of course, 
is to ask that the mixing of $\kappa_\phi$ is uniformly well-behaved over $\Phi$. 
Note that this assumption is not satisfied by many important cases, e.g., a Metropolis--Hastings kernel
based on a Gaussian proposal with a tunable standard deviation $\sigma>0$, which mixes increasingly slowly as $\sigma \to 0$.
In such cases, past work either imposes restrictive assumptions on $\kappa_\phi$, $f$, and $g$ \cite{sun2018mcgd},
or modifies the adaptive algorithm \cref{eq:rm} to operate on a sequence of growing compact subsets of $\Phi$ 
and periodically reset the procedure (e.g.~\cite{chen_truncations_1986,chen_truncations_1988,andrieu_stability_2005,andrieu_expanding_projections_2014}); see \cref{sec:theorydisc} for details.
To avoid the intricacies of either
technique, we  modify the adaptive algorithm to apply $n_t$ kernel updates per step, 
\[
\begin{aligned}
\phi_{t+1} &= \phi_t + \gamma_t g(X_t, \phi_t)\\
X_{t+1} &\dist \kappa_{\phi_{t+1}}^{n_t}(\cdot \given X_t),
\end{aligned}\label{eq:multirm}
\]
and introduce \cref{assump:kernel}, which stipulates 
that $\kappa_\phi$ has uniformly good behaviour on $\Phi$
and $n_t$ increases quickly enough that biases in estimates of $g(\phi)$ 
will not build up over time.
\bassump\label{assump:kernel}
$\sum_{t=0}^\infty \gamma_t^2 < \infty$, and there exist $a, b > 0$ and a sequence $\rho_k \to 0$ such that for all $\phi\in\Phi, x\in\fcX, k \in \nats$,
\[
\max \lt\{
\left| g(\phi)^T \ex g(X^{(\phi,x)}_k, \phi) -  \|g(\phi)\|^2\right|,
\rho_k \ex \lt\|g(X^{(\phi,x)}_k,\phi)\rt\|^2
\rt\}
\leq \rho_k \lt(a + b\|g(\phi)\|^2\rt),
\]
where $X^{(\phi,x)}_k$ is the random variable generated by $k$ applications of the kernel $\kappa_\phi$
to the initial state $X_0 = x$. 
Furthermore, the sequence $n_t$ is designed such that  
\[
\sum_{t=0}^\infty \gamma_t \rho_{n_t} < \infty, \quad \text{ and } \quad 
\sup_t \frac{\gamma_t}{1 - 2\rho_{n_t} a} < \frac{1}{Lb}.
\]
\eassump

\cref{assump:kernel} states that the sequence $n_t$---and hence the computational effort per iteration---must grow
quickly enough such that $\gamma_k \rho_{n_k}$ is summable.
Although this may seem like a heavy price to pay for simplicity, a quickly mixing kernel or a quickly decaying step size leads to very little 
additional work. For example, consider $\gamma_t = 1/(1+t)$, and a uniformly 
geometrically ergodic kernel (over $x$ and $\phi$) with $\rho_k = \rho^k$ for some $\rho\in(0,1)$. 
Then $n_t = \lceil 1+\log(1+t)\rceil$ would satisfy the condition on $n_t$ in \cref{assump:kernel}. 
Note that \cref{assump:kernel} also controls the noise in the process, superceding \cref{assump:boundednoise}
in \cref{thm:markovian} below.

\bthm\label{thm:markovian}
Suppose \cref{assump:stableg,assump:stepsumdiverge,assump:smooth,assump:kernel} hold.
Then 
\[
g(\phi_t) \convp 0, \qquad d(\phi_t, \Phi^\star) \convp 0,  \qquad t\to\infty,
\]
and
\[
\min_{k\in\{0,\dots,t\}}\ex \|g(\phi_k)\|^2 &\leq \frac{f(\phi_0) - \underline f +  \frac{\sum_{k=0}^ta (\gamma_k^2 L + 2\gamma_k \rho_{n_k})}{2}}{\sum_{k=0}^t \gamma_k \lt(1-\frac{2\rho_{n_k} a + \gamma_k b L}{2}\rt)}.
\]

\ethm
\bprf
The proof proceeds identically to that of \cref{thm:deterministic,thm:independentnoise},
except that we apply \cref{assump:kernel} instead of \cref{assump:boundednoise} to control
both $\ex g(\phi_t)^T g(X_t, \phi_t)$ and $\ex \|g(X_t, \phi_t)\|^2$.
\eprf

The final question to answer is whether empirical averages based on the sequence $X_1, \dots, X_T$
converge to expectations under $\pi_{\phi}$. 
For simplicity we assume that $\Phi^\star$ consists of a single point $\Phi^\star = \{\phi^\star\}$,
but similar techniques could be used in the common case where 
$\Phi^\star$ is a set of finitely many local minima of $f$.
\cref{thm:empiricalavg} states that as long as $\pi_{\phi}$ varies continuously in total variation
around $\phi^\star$, and the kernel $\kappa_\phi$ mixes uniformly well near $\phi^\star$, then
empirical averages of the state sequence $X_t$ produced by the multi-draw modified adaptive algorithm \cref{eq:multirm}
behave as expected.

\bassump\label{assump:empiricalavg}
There is a unique solution $\phi^\star\in\Phi$ to the equation $g(\phi) = 0$,
and $\phi_t \to \phi^\star$ implies $\tvd{\pi_{\phi_t}}{\pi_{\phi^\star}} \to 0$.
Further, there exists $\epsilon > 0$ and a sequence $\rho_k \to 0$ such that 
for all $k\in \nats$, distributions $\pi_0$ on $\fcX$, 
and $\phi$ such that $\|\phi - \phi^\star\| \leq \epsilon$,
\[
\tvd{\kappa^k_\phi \pi_0}{\pi_\phi} \leq \rho_k.
\]
\eassump

\bthm\label{thm:empiricalavg}
Suppose \cref{assump:empiricalavg} holds as well as those in \cref{thm:markovian}. Then
for a bounded measurable function $h : \fcX \to \reals$,
\[
\frac{1}{T}\sum_{t=1}^T h(X_t) \convp \ex h(X), \quad X \dist \pi_{\phi^\star}, \qquad T\to\infty.
\]
\ethm
\bprf
For any sequence $(Y_t)_{t=1}^\infty$ of elements in $\fcX$,
\[
\pr\lt(\lt|\frac{1}{T}\sum_{t=1}^T h(X_t) - \frac{1}{T}\sum_{t=1}^T h(Y_t)\rt| > \epsilon\rt)
&\leq 
\pr\lt(\frac{1}{T}\sum_{t=1}^T \lt|h(X_t) - h(Y_t)\rt| > \epsilon\rt)\\
&\leq\pr\lt(\frac{2\|h\|_\infty}{T}\sum_{t=1}^T \ind \lt[X_t \neq Y_t\rt] > \epsilon\rt)\\
&\leq \frac{2\|h\|_\infty}{T\epsilon} \sum_{t=1}^T \pr\lt(X_t \neq Y_t\rt).
\]
Now set $Y_t$ such that conditioned on $\phi_t$, its marginal distribution is $\pi_{\phi_t}$,
but it is maximally coupled with $X_t$. Then
\[
\pr\lt(X_t \neq Y_t\rt) &= \ex \lt[\pr\lt(X_t \neq Y_t \given \phi_t, X_{t-1}\rt)\rt]\\
&= \ex \tvd{\kappa_{\phi_t}^{n_t}}{\pi_{\phi_t}}\\
&= \ex \tvd{\kappa_{\phi_t}^{n_t}}{\pi_{\phi_t}}\ind[|\phi_t -\phi^\star| \leq \epsilon]
+\ex \tvd{\kappa_{\phi_t}^{n_t}}{\pi_{\phi_t}}\ind[|\phi_t -\phi^\star| > \epsilon]\\
&\leq \rho_{n_t} + \pr\lt( |\phi_t - \phi^\star | > \epsilon\rt).
\]
By \cref{assump:empiricalavg} and \cref{thm:markovian}, both terms converge to 0 as $t \to \infty$.
Therefore $\lt|\frac{1}{T}\sum_{t=1}^T h(X_t) - \frac{1}{T}\sum_{t=1}^T h(Y_t)\rt| \convp 0$ as $T\to\infty$.
By the boundedness of $h$ and independence of $(Y_t)_{t=1}^\infty$ given $(\phi_t)_{t=1}^\infty$,
Hoeffding's inequality shows that
\[
\lt|\frac{1}{T}\sum_{t=1}^T h(Y_t) - \frac{1}{T}\sum_{t=1}^T \ex_{\pi_{\phi_t}} h(X)\rt| \convp 0.
\]
Finally, by the definition of total variation distance, 
\[
\lt|\frac{1}{T}\sum_{t=1}^T \ex_{\pi_{\phi_t}} h(X) - \frac{1}{T}\sum_{t=1}^T \ex_{\pi_{\phi^\star}} h(X)\rt|
&\leq \frac{2\|h\|_\infty}{T}\sum_{t=1}^T \tvd{\pi_{\phi_t}}{\pi_{\phi^\star}}.
\]
\cref{thm:markovian} guarantees that $\phi_t \convp \phi^\star$ as $t\to\infty$, 
so $\tvd{\pi_{\phi_t}}{\pi_{\phi^\star}} \convp 0$ as $t\to\infty$ by \cref{assump:empiricalavg}, and hence
the above average converges to 0 in probability. Combining the previous three convergence results with the triangle
inequality yields the stated result.
\eprf

\subsection{Discussion}\label{sec:theorydisc}

In this section, we developed a set of counterexamples to convergence and stability of
the adaptive algorithm \cref{eq:rm} (and its multi-draw modification \cref{eq:multirm}),
which then led to a set of assumptions sufficient to prove convergence in probability of 
$g(\phi_t)$ to 0 and $\phi_t$ to the set of solutions $\{\phi : g(\phi) = 0\}$.
It is worth noting that the results in this section (\cref{thm:deterministic,thm:independentnoise,thm:markovian,thm:empiricalavg}) are based on
techniques from the stochastic optimization literature, which tends to focus 
mostly on settings with independent noise processes $X_t$, and seeks to obtain explicit rates
of convergence in expectation (and hence in probability) of $g(\phi_t)$, $f(\phi_t)$ and $\phi_t$.
There are, however, many other techniques in the broader literature that have been used to study MOI algorithms.

\subsubsection{Confinement methods}

We assumed that various properties of $g$ and
$\kappa_\phi$---such as Lipschitz continuity, mixing properties, etc.---hold
uniformly over the parameter space $\Phi$. As mentioned earlier, neither of
these assumptions holds in a wide array of problems encountered in practice.
One of the simpler approaches to relaxing these assumptions in the literature
involves approximation of the parameter space $\Phi$ by compact subsets
\cite{chen_truncations_1988,chen_truncations_1986}, where the assumptions are
now required to hold only within each compact subset.  In particular,
\cite{andrieu_stability_2005,andrieu_ergodicity_2006} introduce the usage of an
increasing sequence of compact approximations $\Phi_t \uparrow \Phi$ and derive
general conditions under which the sequence $\phi_t$ is guaranteed to remain confined
within one of the compacta almost surely. This literature also tends to obtain stronger
almost sure convergence results, as opposed to the convergence in probability results from this
chapter. Key limitations are the design of the sequence
of compacta and various tuning parameters and sequences, as well as the need to ``reset'' 
the state $\phi_t$ to lie within the first compact restriction $\Phi_0$ in certain circumstances,
leading to poor practical performance. It is possible to avoid these ``resets'' \cite{andrieu_expanding_projections_2014}
at the cost of knowing more about how quickly one can increase the size of the compacta.

\subsubsection{Rates of convergence of $\phi_t$ and empirical averages using $X_t$}

In this chapter, we obtain rates of convergence of $g(\phi_t)$ and a law of large numbers for $X_t$, but
do not prove anything about the rate of convergence of $f(\phi_t)$ and $\phi_t$.
In the stochastic optimization literature, it is typical
to make assumptions about the strict/strong convexity of $f$ to
ensure that there is a unique solution $\phi^\star$ to $g(\phi) = 0$
and to obtain convergence rates for $f(\phi_t)$ and $\phi_t$ \cite{rakhlin_strcvx_2012};
similar techniques should work reasonably for the adaptive algorithm.
See \cite{andrieu_ergodicity_2006} for a much more general treatment of when the states $X_t$
can be used to obtain approximations of expectations.

\subsubsection{Fixed targets and adaptive MCMC}

An important subset of MOI problems that offer some reprieve compared to the general case
is found in adaptive MCMC, where $\pi_\phi = \pi$ does not depend on the parameters,
and \cref{eq:rm} is used to guide improvements to $\kappa_\phi$.
Here the main question of importance is whether the state
sequence $X_t$ provides asymptotically exact estimates of expectations under $\pi$; the behaviour of the parameter
sequence $\phi_t$ itself is not of concern aside from its effect on $X_t$. In this case,
perhaps the simplest way to ensure that the state sequence $X_t$ can be used as approximate draws from $\pi$ 
is to adapt $\kappa_\phi$ increasingly infrequently as time goes on 
(e.g., AirMCMC \cite{chimisov_air_2018}, doubling strategies \cite{surjanovic2022parallel,syed2022nrpt}).
The fixed-target setting also enables another useful strategy: one can stabilize the adaptive kernel $\kappa_\phi$ by 
mixing it with a nonadaptive kernel, $\kappa'_\phi = 0.5\kappa_\phi + 0.5\kappa$, where $\kappa$ is $\pi$-invariant.
This ensures that $\kappa'_\phi$ matches within a factor of 2 the performance of the \emph{better} of $\kappa_\phi$ and $\kappa$,
thus taking advantage of $\kappa_\phi$ when it performs well but falling back to $\kappa$ when
adaptation fails poorly.

\subsubsection{Variance reduction and coreset MCMC}

In general, in stochastic optimization problems, \emph{sublinear} convergence (for example, $\ex f(\phi_t) - \underline f \propto 1/t$)
is usually the best one can hope for, even under strong convexity, due to the presence of noise that constantly perturbs the sequence $\phi_t$ \cite{rakhlin_strcvx_2012}.
However, there is one situation in which faster rates (e.g., $\ex f(\phi_t) - \underline f \propto \exp(-t)$) typical of deterministic first order methods applied to sufficiently nice function (e.g. strongly convex) 
are exhibited by stochastic optimization routines: when the magnitude of the noise decays to 0 as $t\to\infty$. 
This property has been exploited extensively by the literature on \emph{variance-reduced} stochastic optimization
methods; see \cite{gower_variance_2020} for a recent review. Although variance reduction has not been thoroughly explored
yet in the broader MOI literature, it has made an appearance in coreset MCMC \cite{chen2023coreset}. 
Specifically, gradient estimates in coreset MCMC roughly take the form
\[
g(X_t, \phi_t) = C(X_t)(\phi_t - \phi^\star),
\]
for some matrix $C$.
Under certain assumptions, the $\phi_t-\phi^\star$ multiplicative factor in the gradient estimate ensures that
as $\phi_t \to \phi^\star$ the noise in $g(X_t, \phi_t)$ decays, yielding linear convergence \cite[Theorem~3.4]{chen2023coreset}.
Further exploration is warranted to determine whether variance reduction might play a role in MOI problems more generally.

\section{MCMC-driven distribution approximation}
\label{sec:dist_approx}

Having presented the MOI framework, strategies for solving MOI problems, 
and a unified convergence theory for problems falling within this framework, 
we now focus our attention on a detailed example.
In this section we study the use of MCMC for solving the following common MOI problem: 
given a target distribution $\pi$ and a family of approximating distributions 
$\fcQ = \cbra{q_\phi : \phi \in \Phi}$, find a $\phi^\star$ such that 
$q_{\phi^\star} \approx \pi$ with respect to some chosen divergence. 
This $q_{\phi^\star}$ may be of independent interest, but it may also be used 
to assist in further MCMC sampling.

We begin by considering the  problem of learning a proposal for an 
independence Metropolis--Hastings (IMH) kernel. We introduce a  family of 
approximating distributions and show how to minimize the forward KL divergence 
within this family. Then, we discuss tempering as a technique for 
avoiding the curse of dimensionality and tackling challenging high-dimensional 
probability distributions. Finally, we introduce transport map techniques, 
based on methods such as normalizing flows, to further increase the 
expressiveness of $\fcQ$.

\subsection{A simple example: learning via independence kernels}\label{sec:example}
\newcommand{\alpharw}{\alpha_\text{RW}}
\newcommand{\alphaimh}{\alpha_\text{IMH}}

In this example, an independence MH kernel is optimized 
by minimization of the forward KL divergence. 
The focus of this section is on how to formulate the family of distributions 
to be optimized and how to perform the optimization with MCMC.

\subsubsection{A flexible family of approximating distributions}

Denote the vector of $d$ latent variables of interest by 
$X = (X_1, X_2, \dots, X_d)$. For instance, these may be parameters or random effects.
Given $x \in \reals^d$ and an observation $y$, we denote the likelihood by $L(y | x)$. 
Because $y$ is fixed in the Bayesian setup, we omit it from the 
notation and write $L(x) = L(y | x)$. 

For our new family of approximating distributions it is convenient to express the prior as 
$X = G(Z)$ where $Z$ is a (multivariate) standard normal random variable
and $G$ is a deterministic function.
This transformation-based construction is sufficiently general,
capturing many important models including state-space models, spike-and-slap regression, 
and discretizations of diffusion processes.\footnote{For instance, in state-space models 
with $X$ denoting the collection of latent states,
we can write $X_i = G_i(X_{i-1}, Z_i)$ for some non-linear functions $G_i$
encoding the dynamics of an unobserved trajectory. Note that this can be
rewritten to express the entire vector $X = G(Z)$ with a single function $G$. 
Other models can be handled similarly.}
This construction allows us to embed the prior 
into a larger class of distributions, where the other members of 
the class are obtained by going from a standard normal distribution to 
a family of normal distributions. 
We emphasize that we do not make any assumptions on $G$ other than that it can be evaluated pointwise.

\begin{figure}
  \centering
  \includegraphics[width=0.35\linewidth]{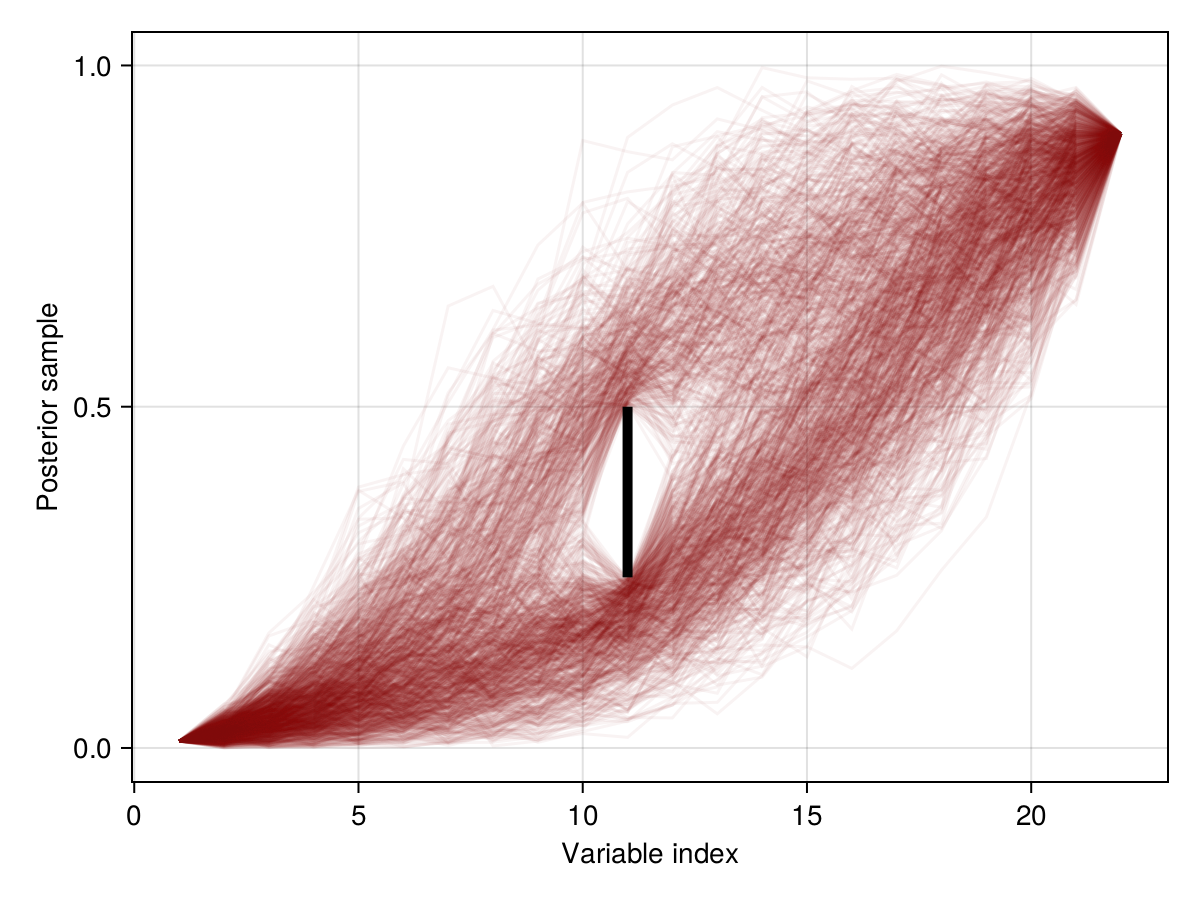}
  \includegraphics[width=0.35\linewidth]{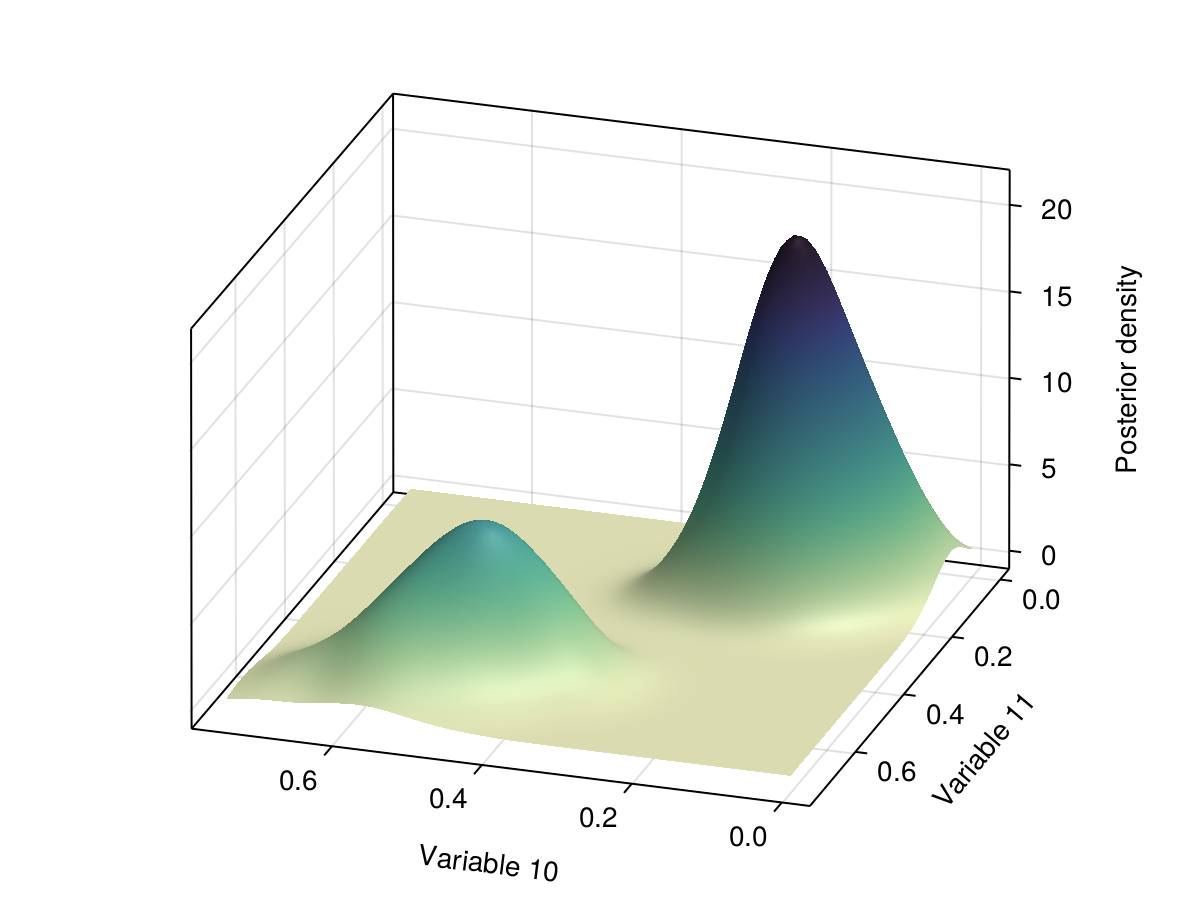}
  \includegraphics[width=0.25\linewidth]{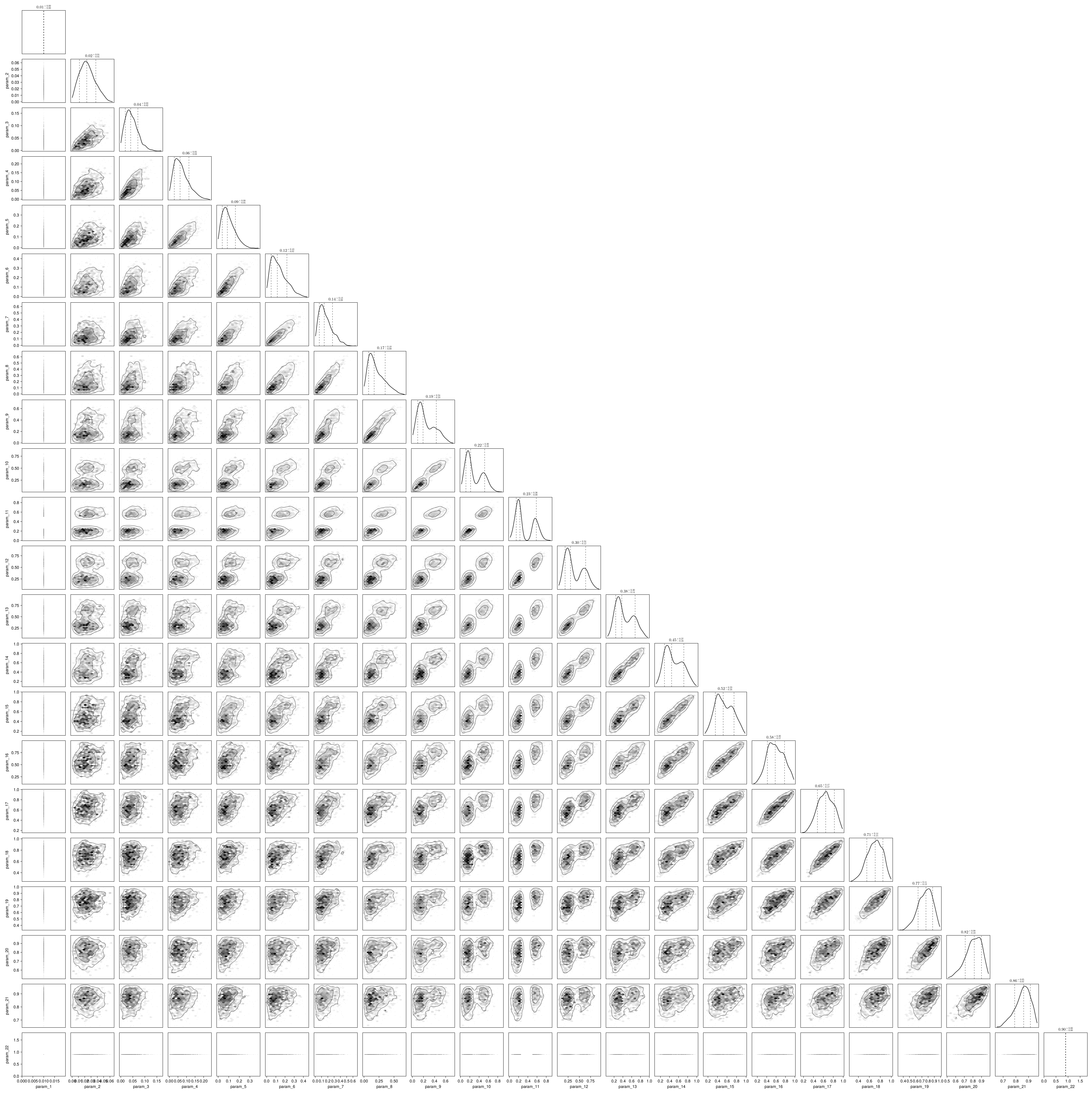}
  \caption{Visualization of the bridge SDE posterior distribution used to 
    illustrate the various algorithms in this section. The SDE is a Wright--Fisher 
    diffusion on $(0, 1)$. 
	\textbf{Left:} Distribution over SDE paths (red) bridging two 
    fixed anchors and avoiding a hard constraint (thick black line). 
    \textbf{Middle:} Pairwise distribution of variables in the neighborhood of the hard 
    constraint showing multimodality. 
	\textbf{Right:} Pair plot of all pairwise posterior distributions.}
  \label{fig:paths}
\end{figure}

\def\proj{\operatornamewithlimits{proj}}

\subsubsection{Running example}

As a concrete running example in this section, 
consider the distribution over paths of a stochastic differential equation (SDE) 
bridging two fixed anchors while avoiding a hard constraint (see \cref{fig:paths}). 
In this setting, we take $X_i$ to be the random value at time point $i$ of the 
discretized process. 
In this example we use a discretization of the Wright--Fisher diffusion with 
Euler--Maruyama step-size $\delta$ \cite{sarkka2019applied}, for which
\[ 
  X_i 
  = G_i(X_{i-1}, Z_i) 
  = \proj_{[0, 1]} \lt( X_{i-1} + \sqrt{ \delta X_{i-1}  (1-X_{i-1}) } Z_i \rt),
\]
where $\proj_{[0,1]}$ denotes a projection onto $[0,1]$ and $Z_i$ is standard normal.
The Wright-Fisher diffusion model is used in population genetics to model the 
evolution of an allele frequency over time \cite{tataru2017statistical}.

\subsubsection{Sampling in $Z$-space}

Because $X$ is a deterministic function of $Z$, one can define a sampler over 
either the $X$ variables (with prior denoted $\pi_0$) or the $Z$ variables 
(with prior denoted $f_0$). 
As we will see soon, in our example it is highly advantageous to define the chain on the variable $Z$, targeting the distribution 
\[  \tilde\pi(z) \propto f_0(z) \; L(G(z)), \]
where $f_0$ denotes the prior distribution on $Z$. 
Doing so will allow us to define a variational family that can be optimized using a simple moment matching algorithm. 
Morevoer, that algorithm achieves the fast convergence of second order methods without 
explictly forming a pre-conditioning matrix. 
Note that $\tilde\pi$ satisfies 
the property that the pushforward of $\tilde\pi$, denoted $\tilde\pi \circ G^{-1}$, 
is equal to $\pi$. In our running Wright-Fisher example, the reparameterized prior $f_0$ is a standard normal distribution.

\subsubsection{Independence kernel Metropolis-Hastings algorithm}\label{eq:imh-review}

Consider one of the simplest possible sampling schemes:
MCMC with an independence Metropolis-Hastings (IMH) kernel.
In this section it will instead
We use IMH as a pedagogical starting point leading
to powerful MOI-based approximations for complex posterior distributions.

In IMH, the proposal $Z_t^*$ does not not depend on $Z_{t-1}$. 
Denote that proposal density by $f_\phi$. We first review the simple situation where 
$\phi$ is fixed, then extend to learning $\phi$ on the fly. 

Recall the transition kernel of IMH is given by:
\[ \label{eq:imh-kernel}
  Z_t \gets \MH_{\alpha_\text{IMH}}(Z_t^* , Z_{t-1}, U_t), \qquad 
  Z_t^* \sim  f_{\phi}, \qquad
  \alphaimh(z^*, z) = 1 \wedge \frac{\tilde\pi(z^*) f_{\phi}(z)}{\tilde\pi(z) f_{\phi}(z^*)},
\]
where $U_t$ is \iid uniform, and the MH kernel $\MH_\alpha(z^*, z, u)$ is equal 
to $z^*$ if $u < \alpha$ and to $z$ otherwise. 

The average rejection rate (RR) of the IMH algorithm at stationarity can be expressed
in terms of the total variation distance. 
That is, for $Z \sim \tilde\pi$ and $Z^* \sim f_\phi$,
\[ \label{eq:tv-bound}
  \text{RR} 
  = 1 - \ex[\alphaimh(Z^*, Z) ] 
  = \tvd{\tilde\pi \times f_\phi}{f_\phi \times \tilde\pi} 
  \leq 2 \tvd{\tilde\pi}{f_\phi}.
\]
We see from \cref{eq:tv-bound} that effective use of IMH relies on the 
construction of a proposal $f_{\phi}$ such that: 
(1) $f_{\phi}$ is close to $\tilde\pi$ in total variation distance; and  
(2) $f_{\phi}$ should be tractable, which we define in our context as being 
able to sample from $f_\phi$ \iid and evaluate the density of $f_\phi$ pointwise.

With \cref{eq:tv-bound} in mind, the choice of the proposal distribution for an IMH algorithm 
can be cast as a minimization problem of the total variation distance 
between $\tilde\pi$ and a proposal distributions $f_\phi$.

\subsubsection{Learning the independence kernel: formulation}

We now formalize an optimization problem aimed at improving the IMH proposal. 
This example draws from the substantial literature on 
optimization of IMH proposals. Early work rooted in variational 
inference \cite{de_freitas_variational_2001} is based on the reverse (exclusive) KL. 
The first implicit instance of forward (inclusive) KL optimization that we are aware of   
is in \cite[Section 3]{gasemyr_adaptive_2003}, and the connection 
to forward KL optimization is made explicit in \cite[Section 7]{andrieu_ergodicity_2006}. 
See also \cite{rue_approximating_2004} and \cite{maire_adaptive_2019} for further developments. 

The space over which we are optimizing consists of a set of candidate proposal 
distributions, which we denote by 
$\fcF = \{ f_\phi : \phi \in \Phi\}$.
In our running example, we take $\fcF$ to be the set of multivariate normal 
distributions with diagonal covariance matrices. 
By construction, $f_0 \in \fcF$, and hence optimization over $\fcF$ should not 
degrade performance compared to naive IMH based on the prior. 
 
With greater generality, we may assume that $\fcF$ is an exponential family, 
which we construct in two steps: first defining  $\bar \fcF$, then reindexing it to obtain $\fcF$. 
First, we define a canonical exponential family, $\bar \fcF = \{\bar f_\eta\}$,
\[ 
  \bar f_\eta(z) = \exp(\eta^\top s(z) - A(\eta)), 
\]
where $s$ is the sufficient statistic, $A$ is the log normalization function, 
and $\eta$ lies in some natural parameter space $E$. 
Second, letting $\sigma$ be the moment mapping,
\[ 
  \sigma(\eta) := \ex[s(Z_\eta)], \quad Z_\eta \sim \bar f_\eta,
\]
the exponential family $\fcF$ under the moment parameterization is given by
\[ 
  \fcF = \{f_\phi : \phi \in \Phi\}, \qquad  
  f_\phi := \bar f_{\sigma^{-1}(\phi)}, \qquad 
  \Phi := \left\{ \sigma(\eta) : \eta \in E \right\}.
\]

While \cref{eq:tv-bound} suggests an objective function based on
the total variation distance,
this objective is seldom directly used as an IMH
tuning objective, with a few exceptions, e.g., \cite[Appendix F.11.2]{surjanovic2022parallel}.
Instead,
the forward KL is often preferred \cite{syed2021parallel,kim2022markov,naesseth2020markovian} in
the MCMC-driven optimization context.
One motivation for this choice is its ease of optimization, as the 
forward KL minimization often admits closed-form updates. 
To see this, 
it is useful to recall the classical connection between forward KL minimization and 
maximum likelihood estimation. Let $\cbra{Z_t}_{t=1}^T$ denote MCMC samples 
with stationary distribution $\pi$. We have that
\[ 
  \argmin_{\phi} \kl{\tilde\pi}{f_\phi} 
  \approx \argmin_{\phi}  \frac{1}{T} \sum_{t=1}^T \log \frac{\tilde\pi(Z_t)}{f_\phi(Z_t)} 
  = \argmax_{\phi} \sum_{t=1}^T \log f_\phi(Z_t),
\]
where we recognize in the final expression the maximum likelihood estimation objective function. 
We can therefore cast the problem of minimizing the forward KL 
as the following ``artificial'' statistical estimation problem: 
treat the samples $Z_t$ as observations, the family $\fcF$ as a statistical 
parametric model, and then perform maximum likelihood estimation from these observations.

Returning to our running example of a Wright-Fisher diffusion, where $\fcF$ consists of 
multivariate normal distributions with diagonal covariance matrices, 
we combine the above connection with the standard fact that the maximum 
likelihood under our Gaussian family has closed-form updates that consist of 
matching empirical first and second moments. 
To generalize this observation to other exponential families, note that, 
for $Z \sim \tilde\pi$,
\[ 
  \argmin_{\eta} \kl{\tilde\pi}{\bar f_\eta} 
  &= \argmax_{\eta} \ex[\log \bar f_\eta(Z)], \\
  &= \argmax_{\eta} \eta\T \ex[s(Z)] - A(\eta). \label{kl-obj}
\]
Taking the gradient with respect to $\eta$, the critical points are given by 
\[ \ex[s(Z)] - \nabla A(\eta) = 0. \]
From basic exponential family properties, 
$\nabla A(\eta) = \ex[s(Z_\eta)] = \sigma(\eta)$. 
Hence, by exponential family convexity, 
\[ 
  \argmin_{\eta} \kl{\tilde\pi}{\bar f_\eta} 
  = \sigma^{-1}( \ex[s(Z)] ).
\]
In the moment parameterization, this is then simply
\[ \label{eq:close-form} 
  \argmin_{\phi} \kl{\tilde\pi}{f_\phi} = \ex[s(Z)].  
\]
In the following section we offer some practical advice for performing this optimization.

\subsubsection{Learning the independence kernel: optimization}

We present algorithms to perform the optimization formulated in the last section. 
In the following, let $Z_1, Z_2, \ldots$ be a sequence from a Markov chain and define 
\[  
  \bar S_T = \frac{1}{T} \sum_{t=1}^T S_t, \qquad 
  S_t =  s\lt( Z_t \rt). 
\]
Suppose that $\ex[\bar S_T] \convas s^\star =  \ex[s(Z)]$, where $Z \sim \tilde\pi$, as $T \to \infty$.
In our context, because the $Z_t$ are Markovian, the convergence above
is guaranteed by the law of large numbers for Markov chains.
We also write $\bar K(\eta) = - \eta\T  s^\star + A(\eta)$ to denote our 
forward KL objective function under the natural parameterization, 
and $K(\phi) = \bar K(\sigma^{-1}(\phi))$ for the objective under the moment parameterization. 

\emph{Naive stochastic gradient:} because it is trivial to obtain stochastic 
gradients with respect to $\bar K$, it is tempting to use a simple 
SGD scheme of the form
\[
  \eta_t \gets \eta_{t-1} - \gamma_t \hat \nabla_t \bar K(\eta_{t-1}),
\]
where the deterministic and stochastic gradients are given by
\[ \label{eq:gradient}
  \nabla \bar K(\eta) = - s^\star + \nabla A(\eta), \qquad 
  \hat \nabla_t \bar K(\eta) = - S_t + \nabla A(\eta).
\] 
However, as we show below, in the present context this choice is strictly dominated in 
terms of performance and complexity of implementation by a 
simple moment-matching scheme \cite[Section 3]{gasemyr_adaptive_2003} that we present below.

\emph{From a pilot MCMC run:} Consider a simplified setup where a pilot or 
warm-up MCMC run has produced estimates of the sufficient 
statistics parameter $\bar S_t$. In that case, working in the 
moment parameterization and using \cref{eq:close-form} yields the update  
$\phi \gets \bar S_t$. 
While the moment parameterization $\phi$ 
is convenient for parameter estimation, it is often not directly supported 
by random number generator and density evaluation libraries, both 
of which are needed when proposing from the learned distribution $f_\phi$.
This means that one has to convert $\phi$ into a more standard parameterization. 
In some cases, such as for normal distributions, this can be done in closed-form. 
In other cases, such as for gamma distributions, the moment mapping 
$\sigma(\eta) = \nabla A(\eta)$ might not have a closed-form inverse. 
When this is the case, one can use a ``mini optimization'' inner loop to compute 
$\sigma^{-1}$. The key point is that this mini-optimization has a compute 
cost independent of the number of samples $t$ and so it is not a 
bottleneck, especially since it can be warm-started at the previous point.
This type of nested optimization is related to mirror descent \cite{amid2020reparameterizing}.

\begin{figure}[t]
	\centering
	 \includegraphics[width=0.6\linewidth]{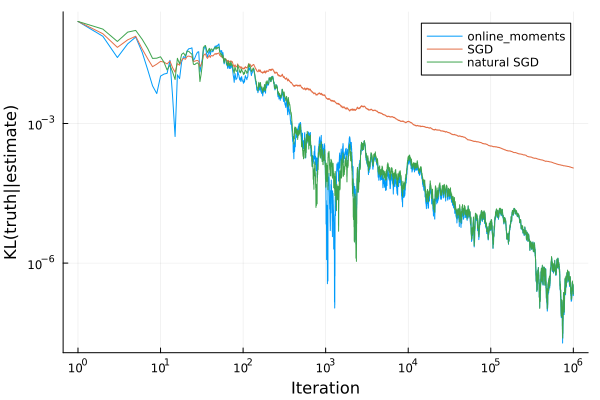}
	\caption{Online learning of a normal distribution with mean $0.2$ and variance $0.4$ from \iid draws. }
	\label{fig:online-normals}
\end{figure}

\emph{Online method:} Instead of doing a pilot MCMC run, one can instead continually update 
$\phi$ using an online mean estimator, such as 
\[ \label{eq:online}
  \phi_t \gets \frac{(t-1) \phi_{t-1} + S_t}{t} = \phi_{t-1} + \frac{S_t - \phi_{t-1}}{t}.
\]
This update can be seen as a special case of the online EM algorithm 
used in \cite[Section 7.4]{andrieu_ergodicity_2006} to learn an 
IMH sampler when setting the number of mixture components to one.
We see in \cref{fig:online-normals} that this online scheme 
strongly outperforms naive SGD.
This might be surprising because the updates superficially appear very similar. 
We demystify this in \cref{sec:moment-as-precond} by recasting the above moment update as a preconditioned 
SGD scheme. 

\emph{Round-based method:} In a distributed computational setting, an online update 
is less attractive as it produces diminishing changes per constant communication 
cost. Instead, one can use a round-based scheme (see \cite{chimisov_air_2018} 
and \cite[Section 5.4]{syed2022nrpt}), based on batches $\bar S^{(r)}$ 
where batch $r$ contains $2^r$ samples based on parameter 
$\phi^{(r)} = \bar S^{(r-1)}$.

Having presented several approaches to performing the optimization, 
we provide an end-to-end MOI algorithm example based on the developed methodology.

\bexa
\label{ex:IMH}
Denote the IMH kernel defined by \cref{eq:imh-kernel} as $\kappa_\phi$. 
At each iteration $t$ we proceed as follows. 
First, we perform one IMH step using the current proposal parameter $\phi_{t-1}$ (\cref{eq:imh-kernel} with $\phi = \phi_{t-1}$):
\[ Z_t \sim \kappa_{\phi_{t-1}}(\cdot | Z_{t-1}). \]
Then, we compute the increment on the sufficient statistics, $S_t \gets s(Z_t)$, 
and update the proposal parameters via \cref{eq:online}.
\eexa

\brmk  \label{sec:moment-as-precond}
To demystify the performance gap between naive SGD and moment matching that can be 
seen in \cref{fig:online-normals}, we review how \cref{eq:online} can be 
interpreted as SGD with a specific preconditioning matrix $P$ and learning rate sequence $\gamma_t$. 
For background on the closely related notion of a natural gradient, see \cite{amari_natural_1998}. 
Recall that pre-conditioned gradient descent of a function $H$ uses updates of the
following form:
\[ \label{eq:generic-newton}
  \phi_t \gets \phi_{t-1} - \gamma_t P(\phi_{t-1})^{-1} \nabla H(\phi_{t-1}). 
\]
For example, Newton's method uses the Hessian matrix as the pre-conditioner, so that
$P(\phi) = \nabla^2  H(\phi)$.

However, to make \cref{eq:generic-newton} coincide 
with moment matching, we have to use a different preconditioner than the Hessian matrix of the objective $H = K$
(recall that $K$ is the forward KL objective 
under the moment parameterization). 
To infer the preconditioner establishing the connection with moment matching, first note that by the chain rule, 
\[  
  \nabla K(\phi) 
  &= \nabla\{ \bar K \circ \sigma^{-1}\}(\phi) \\
  &= \nabla \sigma^{-1}(\phi)\T \nabla\{\bar K\}(\sigma^{-1}(\phi)).
\]
Therefore, if we set $P = (\nabla \sigma^{-1})\T$, we obtain
\[ 
  P(\phi)^{-1} \nabla K(\phi) = \nabla\{\bar K\}(\sigma^{-1}(\phi)). 
\]
Plugging this into \cref{eq:generic-newton} with $H = K$ and $P = (\nabla \sigma^{-1})\T$, 
we obtain the update
\[ 
  \phi_t \gets \phi_{t-1} - \gamma_t \nabla\{\bar K\}(\sigma^{-1}(\phi_{t-1})). 
\]
To conclude our connection to moment matching, we derive a stochastic version of the algorithm. 
From \cref{eq:gradient} and $\nabla A = \sigma$,  
\[
  \hat \nabla_{t} \bar K(\sigma^{-1}(\phi_{t-1})) =  -S_t + \phi_{t-1},
\]
so setting $\gamma_t = 1/t$, we obtain the moment matching update in \cref{eq:online}.

To see that $P$ is distinct from the Hessian of the objective, it is enough to examine the univariate case:
\[
  K'(\phi) &= -(\sigma^{-1})' (s^\star - \phi) \\ 
  K''(\phi) &= -(\sigma^{-1}(\phi))'' (s^\star - \phi) + (\sigma^{-1}(\phi))' \\ 
            &= P(\phi) -(\sigma^{-1}(\phi))'' (s^\star - \phi).
\]
Therefore, $P$ is close to the Hessian $K''$ when $\phi \approx s^\star$.
This behaviour is seen empirically in
\cref{fig:online-normals}: after a sufficient number of iterations, 
online moment matching is eventually indistinguishable from Hessian-conditioned natural gradient.
The take-away message is that
moment matching should be used whenever possible, given its ease of implementation 
and optimality properties \cite[Section 4]{amari_natural_1998}. 
Both methods strongly outperform naive unconditioned SGD. 
\ermk

\begin{figure}
	\centering
	\includegraphics[width=0.6\linewidth]{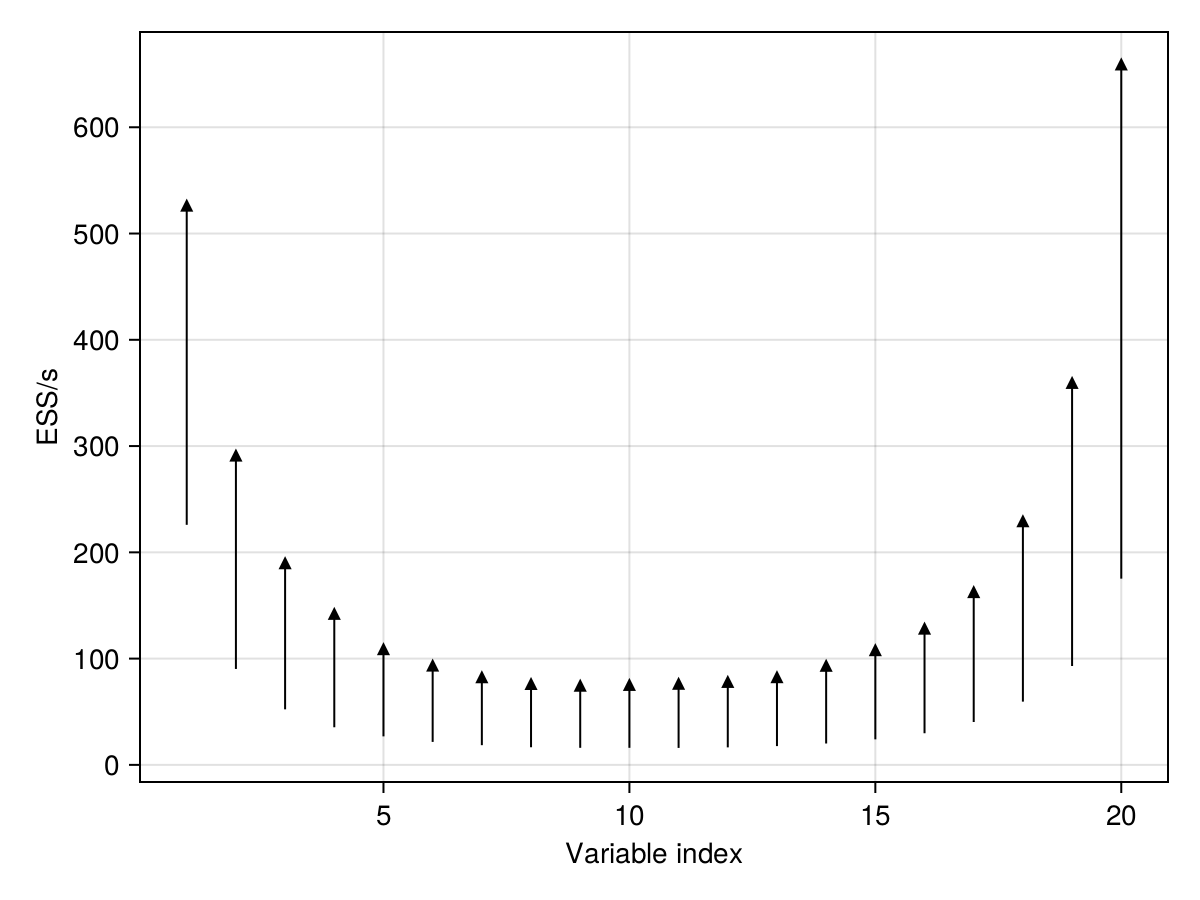}
	\caption{Improvement in ESS per second brought by the optimized IMH algorithm 
    for the constrained Wright--Fisher diffusion.
		For each variable in the model (x-axis), we present the ESS per second 
		(timings include variational training.).
    The baseline is a random walk MH algorithm (point at the beginning of each arrow), 
    which we compare to an optimized IMH algorithm (point at the end of each 
    arrow). Arrows pointing upwards represent an efficiency improvement brought 
    by the optimized IMH kernel.  } 
	\label{fig:imh-ess}
\end{figure}

In \cref{fig:imh-ess} we present the improvements in effective sample size (ESS) 
per second when an optimized IMH kernel is used for the Wright--Fisher diffusion model at 
20 discretized time points. We train the IMH kernel using the approach presented 
in \cref{ex:IMH}, but with the \emph{round-based} tuning procedure describe earlier. 
We alternate one iteration of IMH with one iteration of slice sampling, which performs 
well on targets with varying local geometry even with default settings. 

\subsection{Scalability via tempering}

We previously saw an example of an IMH kernel proposing from $q_\phi$ and targeting $\pi$. 
Such an approach tends to fail in high dimensions due to low MH acceptance probabilities 
as a consequence of the curse of dimensionality, which we illustrate below.
One resolution to this problem is to bridge the gap between $q_\phi$ and $\pi$ 
by introducing a sequence of distributions lying on an \emph{annealing path} between 
the two endpoints. 
We do so using a technique called \emph{parallel tempering}.
The hope is that MH acceptance probabilities between adjacent distributions on the 
path are higher and can help samples from $q_\phi$ travel to $\pi$.
When only two distributions are used on the path, parallel tempering is actually 
equivalent to the IMH method from last section, but increasing the number of 
intermediate distributions in the path is key to high-dimensional scaling. 

\begin{figure}
	\centering
	\includegraphics[width=0.45\linewidth]{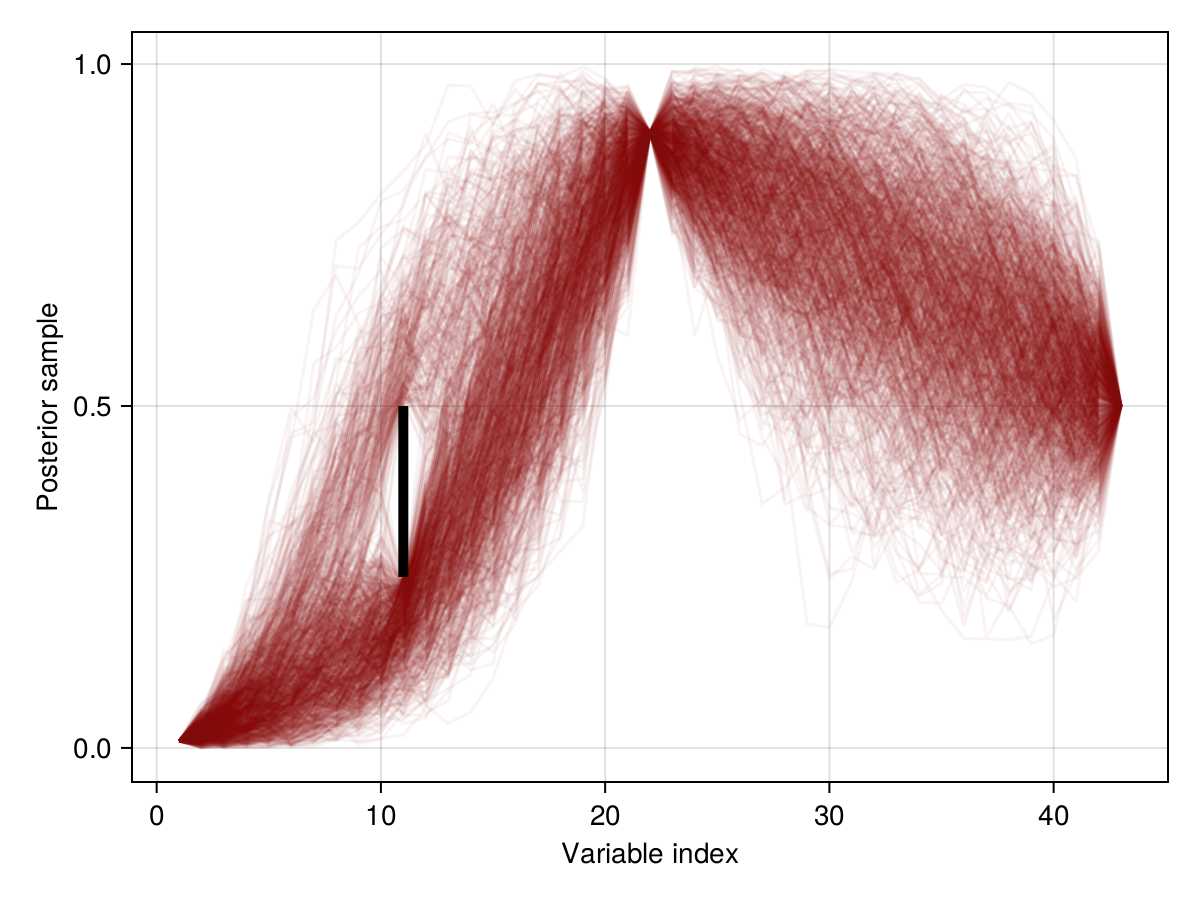}
	\includegraphics[width=0.45\linewidth]{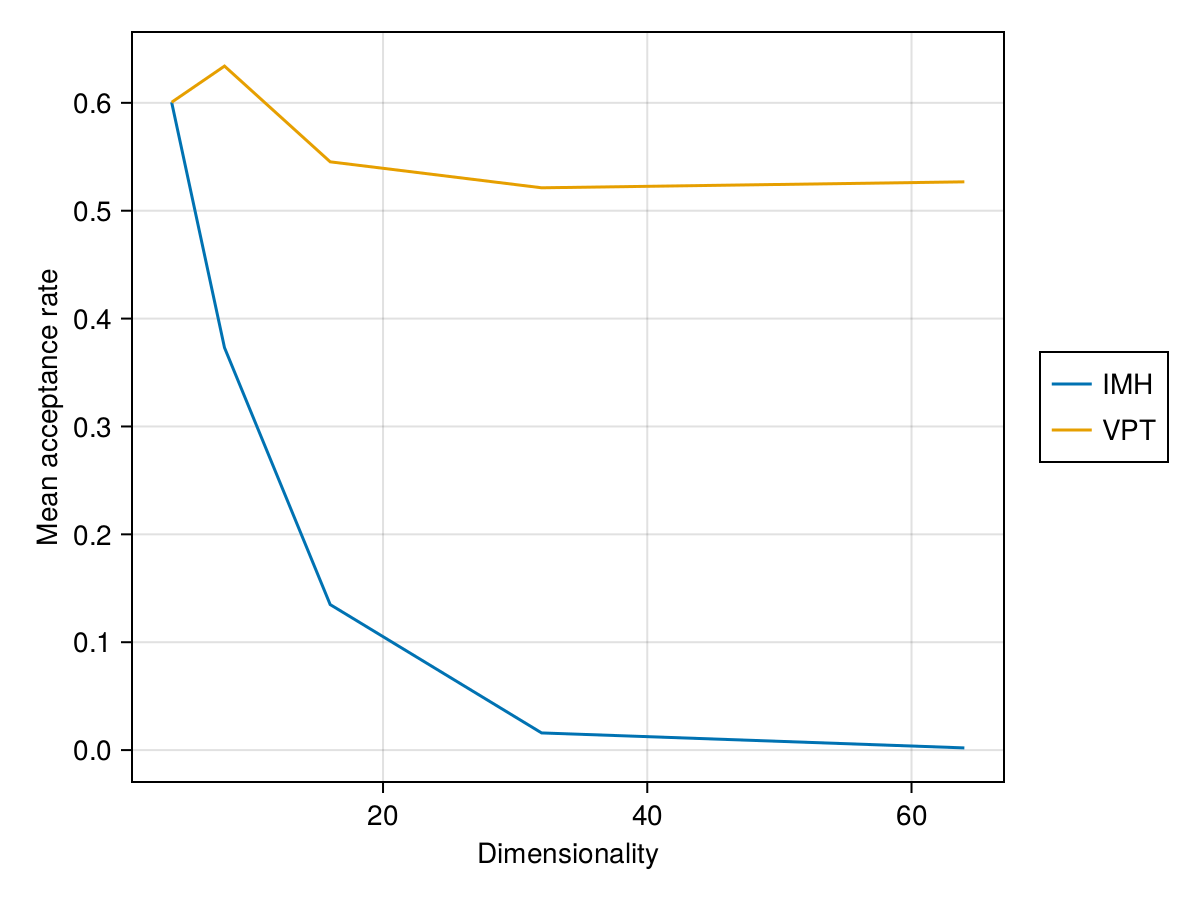}
	\caption{\textbf{Left:} Introduction of more ``observations'' (anchors) in the 
    Wright--Fisher example. 
    Here we show the distribution of paths obtained after adding one additional observation.
		\textbf{Right:} As more observations are introduced, the dimensionality 
    of the posterior distribution increases, and the acceptance rate of the 
    optimized IMH algorithm collapses exponentially fast (blue line). This is 
    addressed by generalizing optimized IMH into a ``variational parallel tempering'' 
    (VPT) algorithm. With VPT, a polynomial increase in the number of PT chains 
    suffices to preserve a constant swap acceptance rate. Specifically, in this example 
    we set the number of chains to $N \propto d^{1/2}$, as predicted in \cite{atchade2011towards}.}
	\label{fig:tempering}
\end{figure}

In the right panel of \cref{fig:tempering} we see that as the dimension of the problem 
increases, the IMH kernel acceptance probability approaches zero. 
To see that IMH proposals can result in very low acceptance probabilities in high dimensions, 
we consider the special case of product distributions.
For $d \geq 1$, consider proposals from $q_\phi^d$ targeting $\pi^d$. Then, 
for $x_0^d, x_1^d \in \fcX^d$, the corresponding IMH acceptance probability is given by 
\[
  \alpha(d, x_0^d, x_1^d) = 1 \wedge 
    \frac{\pi^d(x_1^d) \cdot q_\phi^d(x_0^d)}{\pi^d(x_0^d) \cdot q_\phi^d(x_1^d)}.
\]
It can be shown that at stationarity---i.e., 
$X_0^d \sim \pi^d$ and $X_1^d \sim q_\phi^d$---for any $\eps > 0$ we have
\[
  \alpha(d, X_0^d, X_1^d)^{d^{-1+\eps}} \xrightarrow{p} 0, \qquad d \to \infty.
\]
This suggests that the IMH acceptance probabilities decrease exponentially as 
the dimension of the problem increases and so IMH will generally not perform 
well in high dimensions without an annealing-based approach.

We demonstrate how to use tempering paths to obtain up to an exponential improvement 
for high-dimensional targets, as illustrated in the right panel of \cref{fig:tempering}.
One common path of distributions is given by 
\[ \label{eq:annealing_path}
  \pi_{\phi, \beta} = q_\phi^{1-\beta} \cdot \pi^\beta, \qquad 0 \leq \beta \leq 1.
\]
This simple path, which linearly interpolates between 
log densities and uses a possibly tuned variational reference $q_\phi$ 
from which we can obtain i.i.d.~samples, 
is one of many possible choices \cite{surjanovic2022parallel,syed2022nrpt}. 
This approach can also be seen as a generalization of some previous work on adaptive IMH proposals  
\cite{de_freitas_variational_2001,gasemyr_adaptive_2003,maire_adaptive_2019}.
In the context of PT and annealing methods, this approach is similar to 
\cite[Section 4.1]{cameron_recursive_2014} and \cite[Section 5]{paquet_perturbation_2009}.

In the case of \emph{parallel tempering} (PT), which we describe below, 
one can show that only $O(d)$ samples from $q_\phi^d$ are required to obtain an 
approximately independent sample from $\pi^d$ under appropriate assumptions \cite[Theorems 1; Appendix F, Proposition 4]{syed2022nrpt}, which is in stark contrast to the 
$O(c^d)$ number of samples in the case of non-tempered IMH.
There are several ways to take advantage of the annealing path in \cref{eq:annealing_path} 
to form an effective MCMC algorithm for sampling from $\pi$, including 
parallel tempering (PT) \cite{geyer1991markov,hukushima1996exchange,swendsen1986replica,syed2022nrpt} and 
simulated tempering methods \cite{biron2023nrst,geyer1995annealing}. 
With both methods, a finite number of chains $N$ and an 
annealing schedule $\fcB_N = \cbra{\beta_n}_{n=0}^N$ with 
$0 = \beta_0 \leq \beta_1 \leq \cdots \leq \beta_N = 1$ are selected, 
which induce a sequence of distributions $\pi_{\phi, \beta_n}$ on the annealing path. 
Strategies for selecting the number of chains and an optimal annealing schedule 
exist in the literature on PT \cite{syed2022nrpt}.

In PT, a Markov chain is constructed targeting 
$\pi_{\phi,\beta_0} \times \pi_{\phi,\beta_1} \times \cdots \times \pi_{\phi,\beta_N}$ on the
augmented space $\fcX^{N+1}$. 
Updates on the augmented space proceed by alternating between two different types of Markov kernels 
referred to as \emph{local exploration} and \emph{communication} moves, respectively. 
For a given annealing parameter $\beta_n$, an exploration kernel $K^{(\beta_n)}$ 
is assigned, such as a Hamiltonian Monte Carlo, Gibbs, or slice sampler. 
An exploration move consists of an application of each $K^{(\beta_n)}$ to the $n^\text{th}$ 
marginal of the Markov chain. These updates are non-interacting and can be performed in parallel. 
However, the goal of tempering is to allow samples from the reference $q_\phi$ to travel to the target 
chain corresponding to $\pi$. This is achieved by allowing chains to interact using 
communication moves. These moves allow states between the $n^\text{th}$ and $(n+1)^\text{th}$ 
marginals of the Markov chain to be exchanged with a certain MH acceptance probability.
We present pseudocode for non-reversible parallel tempering \cite{syed2022nrpt} in \cref{alg:nrpt}.
This approach to PT has been shown to provably dominate other PT approaches previously used 
in the literature \cite{syed2022nrpt}. 

\begin{algorithm}[t]
	\footnotesize
\caption{Non-reversible parallel tempering}
\label{alg:nrpt}
\begin{algorithmic}
\Require Initial state $\bar{x}_0 \in \fcX^{N+1}$, annealing schedule $\mathcal{B}_N$, 
\# iterations $T$, annealing path $\pi_\beta$ ($0 \leq \beta \leq 1$)
\For{$t=1,2,\ldots,T$}
  \State $\bar{x} \gets \texttt{LocalExploration}(\bar{x}_{t-1}, \fcB_N, \pi_\beta)$ 
  \Comment{Local exploration kernels}
  \LineComment{$S_\text{even}$ and $S_\text{odd}$ are the even and odd 
    subsets of $\cbra{0, 1, \ldots, N-1}$}
  \State {\bf if } ($t-1$ is even) $S_t \gets S_\text{even}$ {\bf else} $S_t \gets S_\text{odd}$
  \For{$n \in S_t$}
    \State $\alpha \gets 1 \wedge \frac{\pi_{\beta_n}(x^{n+1}) \cdot \pi_{\beta_{n+1}}(x^n)}
      {\pi_{\beta_n}(x^n) \cdot \pi_{\beta_{n+1}}(x^{n+1})}$ 
    \State $U \gets \text{Unif}(0,1)$
    \State {\bf if} {$U \leq \alpha$} {\bf then}  
      $x^{n+1}, \, x^n \gets x^{n}, \, x^{n+1}$ 
      \Comment{Swap components $n$ and $n+1$ of $\bar{x}$}
  \EndFor
  \State $\bar{x}_t\gets \bar{x}$
\EndFor
\State \textbf{Return:} $(\bar{x}_t)_{t=0}^T$
\end{algorithmic}
\end{algorithm}

It is preferable to parallelize computation 
because of the demands of working on the considerably larger augmented space $\fcX^{N+1}$. 
An efficient distributed implementation of PT in the Julia programming language 
is provided by \cite{surjanovic2023pigeons}; this software is also used throughout 
this chapter to demonstrate several useful examples.

\subsubsection{Stabilization}
\label{sec:variational_pt}

In our exposition of PT above, we introduced an annealing path between a variational reference 
$q_\phi$ and fixed target $\pi$, given by \cref{eq:annealing_path}. 
This presentation has an added degree of freedom due to the ability to tune the reference;
traditionally, a tempering path between a fixed reference $\pi_0$ and target  
$\pi$ is used. 
However, stable estimates of $\phi$ are necessary in this case:
a locally (but not globally) optimal choice of $\phi$ can lead to poor samples 
that further reinforce our selection of $\phi$ in a non-global optimum. 
This issue is predicted by PT theory \cite{surjanovic2022parallel}.

In the case of IMH, this issue can perhaps be seen even more clearly. 
If the IMH proposal from $q_\phi$ learns to only propose within one mode of a distribution, 
then we will see mode collapse: proposed samples will be from the given mode and it 
will be difficult to learn a better global proposal for $q_\phi$.
In this case, one simple solution might be to alternate between the proposal $q_\phi$ 
and some fixed proposal $\pi_0$ to help avoid this issue of mode collapse.

To circumvent this issue in PT, it is necessary to stabilize PT when a variational reference 
$q_\phi$ is used. One solution is to introduce multiple reference distributions, 
by introducing multiple ``legs'' of PT \cite{surjanovic2022parallel}. 
For instance, one possible annealing path is 
\[
  \pi_{\phi, \beta} \propto 
  \begin{cases}
    q_\phi^{1-2\beta} \cdot \pi^{2\beta},  & 0 \leq \beta \leq 0.5 \\
    \pi_0^{2\beta-1} \cdot \pi^{2-2\beta}, & 0.5 < \beta \leq 1 \\
  \end{cases}  
\]
For more details about stabilizing PT using multiple reference distributions, 
we refer readers to \cite{surjanovic2022parallel}.

\subsection{Improving expressiveness using approximate transport}
\label{sec:transport} 
In this section we demonstrate one way of further increasing the expressiveness 
of $q_\phi$ using approximate transport maps. 
Transport-based methods construct a map, $f_\phi : \fcX \to \fcX$ for some $\phi \in \Phi$, 
such that the pushforward measure of some simple distribution $q$ with respect to 
$f_\phi$, denoted $q \circ f_\phi^{-1}$, is close to the target distribution $\pi$. 
In almost all cases $f_\phi$ is a diffeomorphism (both the map and its inverse 
are differentiable) for all $\phi$ so that the distribution of $q_\phi \equiv q \circ f_\phi^{-1}$ 
can be obtained by the change of variables formula. 
To find an $f_\phi$ such that $q_\phi$ is close to $\pi$, one typically 
minimizes $\kl{q_\phi}{\pi}$.
In the idealized setting, $q$ is a distribution from 
which we can obtain i.i.d.~samples, such as a multivariate Gaussian, and hence 
minimization of the (reverse) KL divergence above can be achieved with gradient descent.

With the recent popularization of transport-based methods for inference such as 
normalizing flows (NFs) \cite{dinh2016density,durkan2019neural,papamakarios2021normalizing}, 
a natural question that arises is whether these powerful methods 
can be applied to MCMC to assist MCMC exploration. Additionally, MCMC methods 
might be useful for minimizing other divergences such as the forward (inclusive) 
KL divergence $\kl{\pi}{q_\phi}$.
A review of some advantages and drawbacks of methods that combine 
MCMC and normalizing flow methods is given by \cite{grenioux2023sampling}.

One approach, that we elaborate on in this section, is to (repeatedly) use a transport map from $q$ to 
$\pi$ and then use MCMC to sample from a distribution close to $\pi$
\cite{arbel2021aft,gabrie2022adaptive, hagemann2022snf, matthews2022craft, samsonov2022local, wu2020snf}. 
We illustrate one example of such a method below, based on the work of \cite{gabrie2022adaptive}. 
We offer more details on other uses of approximate transport maps in conjunction with MCMC 
later in this section.

Given a transport map $f_\phi$ such that $q_\phi = q \circ f_\phi^{-1} \approx \pi$, and a 
$\pi$-invariant Markov kernel $\kappa$, 
we define the transport-based IMH kernel $\kappa_T$ as 
\[
  \kappa_T(x, \dee y; \phi) 
    &= \alpha(x, y) q_\phi(\dee y) + (1-r(x)) \delta_x(\dee y) \\ 
  r(x) &= \int \alpha(x, y) q_\phi(\dee y) \\
  \alpha(x, y) &= 1 \wedge \frac{\pi(y) q_\phi(x)}{\pi(x) q_\phi(y)}.  
\]
We can then compose $\kappa$ and $\kappa_T$ to obtain a composite kernel targeting $\pi$.

Following the work of \cite{gabrie2022adaptive}, we update the parameters $\phi$ 
of the transport map as follows. 
The algorithm is presented in \cref{alg:adaptive_transport}.
We start with $N$ particles, preferably initialized within each of the separate modes of the 
target distribution $\pi$. The states of these particles are denoted $\cbra{x_n}_{n=1}^N$.
These particles are then separately updated using $a \geq 1$ applications of the $\pi$-invariant 
Markov kernel $\kappa$, followed by one application of the IMH transport map kernel $\kappa_T$.
Along the way, obtained samples $\cbra{x_n(k)}_{n=1}^N$ for $k \geq 1$ are used to 
update the transport map $f_\phi$. 
Instead of minimizing the forward KL divergence $\kl{\pi}{q_\phi}$, which requires 
samples from the intractable target $\pi$, the approach taken in \cite{gabrie2022adaptive} 
is to minimize a KL divergence between $q_\phi \kappa^k$ and $q_\phi$. Specifically, 
at time step $k$,
the loss for a given $\phi$ is proportional to $\kl{q_\phi \kappa^k}{q_\phi}$ 
and the estimate of the loss is given by 
\[
  \fcL_N(\phi) = -\frac{1}{N} \sum_{n=1}^N \log q_\phi(x_n(k)).
\]
This loss is then minimized using gradient descent.
The output of the algorithm is an approximating distribution $q_\phi$, as well 
as a collection of samples approximately distributed according to $\pi$, 
$\cbra{x_n(k)}_{k=0,n=1}^{k_\text{max},N}$.

\begin{algorithm}[t]
	\footnotesize
\caption{Adaptive MC with transport maps \cite{gabrie2022adaptive}}
\label{alg:adaptive_transport}
\begin{algorithmic}
\Require target $\pi$, initial map $f_\phi$, starting particles $\cbra{x_n(0)}_{n=1}^N$, 
total duration $k_\text{max}$, number of applications of $\kappa$ before applying $f_\phi$: $a$, 
gradient step size $\eps$
\For{$k=0,2,\ldots,k_\text{max}-1$}
  \For{$n=1,2,\ldots,N$} 
    \If{$k \text{ mod } a+1 = 0$}
      \State $x_n(k+1) \sim \kappa_T(x_n(k), \cdot)$ 
    \Else 
      \State $x_n(k+1) \sim \kappa(x_n(k), \cdot)$ 
    \EndIf
  \EndFor
  \State $\fcL_N(\phi) \gets -\frac{1}{N} \sum_{n=1}^N \log q_\phi(x_n(k+1))$
  \State $\phi \gets \phi - \eps \nabla_\phi \fcL_N(\phi)$
\EndFor
\State \textbf{Return:} $\cbra{x_n(k)}_{k=0,n=1}^{k_\text{max},N}, f_\phi$
\end{algorithmic}
\end{algorithm}

\cref{alg:adaptive_transport} can be connected to the variational PT algorithm 
presented in \cref{sec:variational_pt}. In \cite{gabrie2022adaptive}, the authors 
remark that the use of a fixed kernel $\kappa$ in conjunction with the adaptive $\kappa_T$ 
``improves the robustness of the scheme by ensuring that sampling proceeds in 
places within the modes where the map is not optimal,'' which is similar to the 
variational PT stabilization procedure previously mentioned. 
However, referring to \cref{alg:adaptive_transport}, 
``one should not expect the procedure to find states in basins distinct from initialization''
\cite{gabrie2022adaptive}.
This is because the fixed kernel $\kappa$ will likely remain trapped within a given 
mode of initialization; such problems can be avoided with the use of annealing.
There is therefore likely room for annealing-based methods to be used to improve \cref{alg:adaptive_transport}.

\subsubsection{A note on approximate transport maps for pre-conditioning}

An alternative method of combining MCMC and transport, compared to the method introduced 
above, is to construct a map $f_\phi$, sample from $\pi \circ f_\phi$ using MCMC, 
and then push forward the obtained samples through $f_\phi$ 
\cite{cabezas2023transport, hoffman2019neutra, marzouk2016introduction, parno2018transport}.
The main premise is that $\pi \circ f_\phi$ should lie close to $q$, which is 
chosen to be a distribution that is easy to sample from.
These methods may work well when both $q$ and $\pi$ are log-concave distributions, 
however if the distributions have a different structure (e.g., $\pi$ is multimodal), 
then it can be difficult to find an invertible transform $f_\phi$ that can 
capture this structure. 

The use of such transport maps for the purposes of performing MCMC sampling on a 
simpler space was studied by \cite[Section 2.4]{meng_warp_2002},  
\cite{marzouk2016introduction} and \cite{parno2018transport}. 
For example, \cite{marzouk2016introduction} 
consider a polynomial-based approach for $f_\phi$. 
However, this approach was noted by \cite{hoffman2019neutra} to be too expensive 
when the target distribution lives in a high-dimensional space.
To this end, \cite{hoffman2019neutra} consider using inverse autoregressive flows 
(IAFs) and Hamiltonian Monte Carlo (HMC) on the $q$ space to improve the performance of the method.  
In a similar vein, \cite{cabezas2023transport} consider training a map $f_\phi$ 
to minimize $\kl{q \circ f_\phi^{-1}}{\pi}$ where $q$ is chosen to be a 
multivariate normal distribution. Then, an elliptical slice sampler 
\cite{murray2010elliptical} is used to sample from $\pi \circ f_\phi$, which 
should be approximately normally distributed.

\section{Discussion}
\label{sec:discussion} 

In this chapter we presented a unifying framework for many problems at the intersection 
of MCMC and machine learning, which we refer to as Markovian optimization-integration 
(MOI) problems. 
These problems encompass black-box variational inference, adaptive MCMC,
normalizing flow construction and transport-assisted MCMC, surrogate-likelihood MCMC,
coreset construction for MCMC with big data, Markov chain gradient descent,
Markovian score climbing, and more. 
By unifying these problems, theory and methods developed for each may be 
translated and generalized. After presenting a list of some common MOI problems, 
we presented an example with techniques for MCMC-driven distribution approximation, 
as well as some other common strategies for tackling MOI problems and some convergence theory.

Although the MOI framework is very general, there are still some optimization-integration 
problems that cannot be captured with this approach. 
For instance, it may be the case that the Markovian assumption presented is still too strong. 
To achieve an even greater level of generality, we may wish to substitute results 
for gradient estimates with Markovian noise with any estimators that are 
asymptotically consistent. However, as we discussed in \cref{sec:theory}, 
replacing an \iid assumption with a Markovian noise assumption leads to a more complicated 
set of sufficient theoretical conditions that guarantee convergence; generalizing beyond a 
Markovian noise assumption could yield even greater difficulties for the 
study of theoretical convergence properties.

\section{Further reading}
\label{sec:further_reading}

\subsection{In this book}

Chapter~2 discusses the 
adaptive MCMC literature, which is complementary to the present chapter. 
In the setups explored in the chapter on adaptive MCMC, the optimized parameter 
$\phi^\star$ is in itself not of primary interest; for instance, $\phi$ may be a 
proposal bandwidth in that setting. In contrast, for the class of methods 
reviewed here, finding $\phi^\star$ is often a part of or even the whole objective. 
For instance, one may consider the case where $\phi$ parameterizes a rich 
parametric family approximating the posterior distribution. 

Chapter~23 covers state-of-the-art methods to perform
Bayesian inference over deep neural networks, while the methods explored in the present chapter are agnostic to the nature of the posterior distribution. 
Deep neural networks are often part of MCMC-driven learning methods, but focus so far has  been on fitting them using non-Bayesian methods.

\subsection{Other connections}

``Big data'' has been a major focus in both machine learning and statistics in 
the past 10 years. 
\cref{sec:examples} reviews many important techniques used to 
approach large datasets, including black-box variational inference 
and coreset MCMC, both of which use as a building block the 
mini-batch gradient estimation reviewed in \cref{sec:mini-batch}.
However, this only scratches the surface of the range of methods 
developed to approach large datasets. 
See, for example, \cite{angelino2016patterns} for a review of other 
techniques such as data subsampling and divide-and-conquer MCMC methods. 

While we have focused mostly on the machine learning, statistics and optimization 
literatures, MOI problems have been studied in many other fields. 
In particular, they arise in reinforcement learning and control under the name 
of ``average reward problems''   \cite{mahadevan_average_1996}.
Even within a given field, it is often studied separately in different subfields. 
For example, in machine learning they are studied both in the optic of large-scale 
training (where gradient stochasticity is a purely algorithmic device) and in the 
sub-field of online learning (where the stochasiticity is viewed as epistemic) 
\cite{cesa-bianchi_prediction_2006}. 
Similarly, due to space constraints we have skipped other important MOI applications, 
for example to Monte Carlo and stochastic EM \cite{fort_convergence_2003,ruth_review_2024}.

For simplicity, we have assumed throughout the chapter that  
the Markov kernel $\kappa_\phi$ is exactly $\pi_\phi$-invariant. 
However, there has been recently a resurgence of interest in methods 
where  $\kappa_\phi$ is only approximately $\pi_\phi$-invariant 
 \cite{durmus_high-dimensional_2019,welling_bayesian_2011}. 
The most popular example comes from unadjusted approximations 
to Langevin dynamics \cite{ermak_computer_1975,neal_bayesian_1992}, 
often combined with stochastic gradients. 
These ``unadjusted'' methods typically have a worse asymptotic rate of convergence 
compared to exactly invariant kernels. Instead, unadjusted methods 
are typically advantageous in the initial ``burn-in'' phase \cite{teh_consistency_2016},
making them a good match with the MOI setup.

\bibliographystyle{plain}
\bibliography{main}

\begin{thebibliography}{100}

\bibitem{agarwal_information-theoretic_2012}
Alekh Agarwal, Peter~L. Bartlett, Pradeep Ravikumar, and Martin~J. Wainwright.
\newblock Information-theoretic lower bounds on the oracle complexity of
  stochastic convex optimization.
\newblock {\em IEEE Transactions on Information Theory}, 58(5):3235--3249,
  2012.

\bibitem{amari_natural_1998}
Shun'ichi Amari.
\newblock Natural gradient works efficiently in learning.
\newblock {\em Neural Computation}, 10(2):251--276, 1998.

\bibitem{amid2020reparameterizing}
Ehsan Amid and Manfred~K. Warmuth.
\newblock Reparameterizing mirror descent as gradient descent.
\newblock {\em Advances in Neural Information Processing Systems},
  33:8430--8439, 2020.

\bibitem{andrieu_stability_2005}
Christophe Andrieu, \'{E}ric Moulines, and Pierre Priouret.
\newblock Stability of stochastic approximation under verifiable conditions.
\newblock In {\em IEEE Conference on Decision and Control}, 2005.

\bibitem{andrieu_ergodicity_2006}
Christophe Andrieu and \'{E}ric Moulines.
\newblock On the ergodicity properties of some adaptive {MCMC} algorithms.
\newblock {\em The Annals of Applied Probability}, 16(3):1462--1505, 2006.

\bibitem{andrieu_controlled_mcmc_2001}
Christophe Andrieu and Christian Robert.
\newblock Controlled {MCMC} for optimal sampling.
\newblock {\em Cahiers du C\'{e}r\'{e}made 0125}, 2001.

\bibitem{andrieu2008tutorial}
Christophe Andrieu and Johannes Thoms.
\newblock A tutorial on adaptive {MCMC}.
\newblock {\em Statistics and Computing}, 18:343--373, 2008.

\bibitem{andrieu_expanding_projections_2014}
Christophe Andrieu and Matti Vihola.
\newblock Markovian stochastic approximation with expanding projections.
\newblock {\em Bernoulli}, 20(2):545--585, 2014.

\bibitem{angelino2016patterns}
Elaine Angelino, Matthew~James Johnson, Ryan~P Adams, et~al.
\newblock Patterns of scalable {B}ayesian inference.
\newblock {\em Foundations and Trends in Machine Learning}, 9(2-3):119--247,
  2016.

\bibitem{arbel2021aft}
Michael Arbel, Alex Matthews, and Arnaud Doucet.
\newblock Annealed flow transport {M}onte {C}arlo.
\newblock In {\em International Conference on Machine Learning}, pages
  318--330. PMLR, 2021.

\bibitem{atchade2011towards}
Yves~F Atchad{\'e}, Gareth~O Roberts, and Jeffrey~S Rosenthal.
\newblock Towards optimal scaling of {M}etropolis-coupled {M}arkov chain
  {M}onte {C}arlo.
\newblock {\em Statistics and Computing}, 21:555--568, 2011.

\bibitem{baydin2018autodiff}
Atilim~Gunes Baydin, Barak~A Pearlmutter, Alexey~Andreyevich Radul, and
  Jeffrey~Mark Siskind.
\newblock Automatic differentiation in machine learning: {A} survey.
\newblock {\em Journal of Machine Learning Research}, 18:1--43, 2018.

\bibitem{beaumont2002approximate}
Mark~A Beaumont, Wenyang Zhang, and David~J Balding.
\newblock Approximate {B}ayesian computation in population genetics.
\newblock {\em Genetics}, 162(4):2025--2035, 2002.

\bibitem{benveniste2012adaptive}
Albert Benveniste, Michel M{\'e}tivier, and Pierre Priouret.
\newblock {\em Adaptive Algorithms and Stochastic Approximations}, volume~22.
\newblock Springer Science \& Business Media, 2012.

\bibitem{biron2023nrst}
Miguel Biron-Lattes, Trevor Campbell, and Alexandre Bouchard-C{\^o}t{\'e}.
\newblock Automatic regenerative simulation via non-reversible simulated
  tempering.
\newblock {\em arXiv:2309.05578}, 2023.

\bibitem{blei2017variational}
David~M Blei, Alp Kucukelbir, and Jon~D McAuliffe.
\newblock Variational inference: {A} review for statisticians.
\newblock {\em Journal of the American Statistical Association},
  112(518):859--877, 2017.

\bibitem{bordes_sgd-qn_2009}
Antoine Bordes, Léon Bottou, and Patrick Gallinari.
\newblock {SGD}-{QN}: {Careful} quasi-{Newton} stochastic gradient descent.
\newblock {\em Journal of Machine Learning Research}, 10(59):1737--1754, 2009.

\bibitem{jax2018github}
James Bradbury, Roy Frostig, Peter Hawkins, Matthew~James Johnson, Chris Leary,
  Dougal Maclaurin, George Necula, Adam Paszke, Jake Vander{P}las, Skye
  Wanderman-{M}ilne, and Qiao Zhang.
\newblock {JAX}: {C}omposable transformations of {P}ython+{N}um{P}y programs,
  2018.

\bibitem{cabezas2023transport}
Alberto Cabezas and Christopher Nemeth.
\newblock Transport elliptical slice sampling.
\newblock In {\em Artificial Intelligence and Statistics}, pages 3664--3676.
  PMLR, 2023.

\bibitem{cameron_recursive_2014}
Ewan Cameron and Anthony Pettitt.
\newblock Recursive pathways to marginal likelihood estimation with
  prior-sensitivity analysis.
\newblock {\em Statistical Science}, 29(3):397--419, 2014.

\bibitem{campbell2019sparse}
Trevor Campbell and Boyan Beronov.
\newblock Sparse variational inference: Bayesian coresets from scratch.
\newblock {\em Advances in Neural Information Processing Systems}, 32, 2019.

\bibitem{campbell2018bayesian}
Trevor Campbell and Tamara Broderick.
\newblock Bayesian coreset construction via greedy iterative geodesic ascent.
\newblock In {\em International Conference on Machine Learning}, pages
  698--706. PMLR, 2018.

\bibitem{campbell2019automated}
Trevor Campbell and Tamara Broderick.
\newblock Automated scalable {B}ayesian inference via {H}ilbert coresets.
\newblock {\em The Journal of Machine Learning Research}, 20(1):551--588, 2019.

\bibitem{carmon_making_2022}
Yair Carmon and Oliver Hinder.
\newblock Making {SGD} parameter-free.
\newblock In {\em Proceedings of {Thirty} {Fifth} {Conference} on {Learning}
  {Theory}}, pages 2360--2389. PMLR, 2022.

\bibitem{casella1996rao}
George Casella and Christian~P Robert.
\newblock Rao-{B}lackwellisation of sampling schemes.
\newblock {\em Biometrika}, 83(1):81--94, 1996.

\bibitem{cesa-bianchi_prediction_2006}
Nicolò Cesa-Bianchi and Gábor Lugosi.
\newblock {\em Prediction, Learning, and Games}.
\newblock Cambridge University Press, Cambridge ; New York, 2006.

\bibitem{chen_truncations_1988}
Han-Fu Chen, Lei Guo, and Ai-Jun Gao.
\newblock Convergence and robustness of the {R}obbins--{M}onro algorithm
  truncated at randomly varying bounds.
\newblock {\em Stochastic Processes and their Applications}, 27:217--231, 1988.

\bibitem{chen_truncations_1986}
Han-Fu Chen and Yun-Min Zhu.
\newblock Stochastic approximation procedures with randomly varying
  truncations.
\newblock {\em Scientia Sinica 1}, 29:914--926, 1986.

\bibitem{chen2023coreset}
Naitong Chen and Trevor Campbell.
\newblock Coreset {M}arkov chain {M}onte {C}arlo.
\newblock {\em arXiv:2310.17063}, 2023.

\bibitem{chen2022bayesian}
Naitong Chen, Zuheng Xu, and Trevor Campbell.
\newblock Bayesian inference via sparse {H}amiltonian flows.
\newblock {\em Advances in Neural Information Processing Systems},
  35:20876--20888, 2022.

\bibitem{chimisov_air_2018}
Cyril Chimisov, Krzysztof Latuszynski, and Gareth Roberts.
\newblock Air {Markov} chain {Monte} {Carlo}, 2018.
\newblock arXiv:1801.09309.

\bibitem{de_freitas_variational_2001}
Nando de~Freitas, Pedro Højen-Sørensen, Michael~I Jordan, and Stuart Russell.
\newblock Variational {MCMC}.
\newblock In {\em Proceedings of the {Seventeenth} {C}onference on
  {Uncertainty} in {A}rtificial {I}ntelligence}, pages 120--127, 2001.

\bibitem{dembo_truncated-newton_1983}
Ron~S. Dembo and Trond Steihaug.
\newblock Truncated-{Newton} algorithms for large-scale unconstrained
  optimization.
\newblock {\em Mathematical Programming}, 26(2):190--212, 1983.

\bibitem{dinh2016density}
Laurent Dinh, Jascha Sohl-Dickstein, and Samy Bengio.
\newblock Density estimation using {R}eal {NVP}.
\newblock {\em arXiv:1605.08803}, 2016.

\bibitem{duchi2011adaptive}
John Duchi, Elad Hazan, and Yoram Singer.
\newblock Adaptive subgradient methods for online learning and stochastic
  optimization.
\newblock {\em Journal of Machine Learning Research}, 12(7), 2011.

\bibitem{durkan2019neural}
Conor Durkan, Artur Bekasov, Iain Murray, and George Papamakarios.
\newblock Neural spline flows.
\newblock {\em Advances in Neural Information Processing Systems}, 32, 2019.

\bibitem{durmus_high-dimensional_2019}
Alain Durmus and {\'{E}}ric Moulines.
\newblock High-dimensional {Bayesian} inference via the unadjusted {Langevin}
  algorithm.
\newblock {\em Bernoulli}, 25(4A), 2019.

\bibitem{durmus_finite-time_2023}
Alain Durmus, {\'{E}}ric Moulines, Alexey Naumov, and Sergey Samsonov.
\newblock Finite-time {High}-probability {Bounds} for {Polyak}-{Ruppert}
  {Averaged} {Iterates} of {Linear} {Stochastic} {Approximation}, 2023.
\newblock arXiv:2207.04475.

\bibitem{ermak_computer_1975}
Donald~L. Ermak.
\newblock A computer simulation of charged particles in solution.
\newblock {\em The Journal of Chemical Physics}, 62(10):4189--4196, 1975.

\bibitem{fort_convergence_2003}
Gersende Fort and {\'{E}}ric Moulines.
\newblock Convergence of the {Monte} {Carlo} expectation maximization for
  curved exponential families.
\newblock {\em The Annals of Statistics}, 31(4):1220--1259, 2003.

\bibitem{friedlander_hybrid_2012}
Michael~P. Friedlander and Mark Schmidt.
\newblock Hybrid deterministic-stochastic methods for data fitting.
\newblock {\em SIAM Journal on Scientific Computing}, 34(3):A1380--A1405, 2012.

\bibitem{gabrie2022adaptive}
Marylou Gabri{\'e}, Grant~M Rotskoff, and Eric Vanden-Eijnden.
\newblock Adaptive {M}onte {C}arlo augmented with normalizing flows.
\newblock {\em Proceedings of the National Academy of Sciences},
  119(10):e2109420119, 2022.

\bibitem{gasemyr_adaptive_2003}
J{\o}rund G{\aa}semyr.
\newblock On an adaptive version of the {M}etropolis--{H}astings algorithm with
  independent proposal distribution.
\newblock {\em Scandinavian Journal of Statistics}, 30(1):159--173, 2003.

\bibitem{Geffner21}
Tomas Geffner and Justin Domke.
\newblock {MCMC} variational inference via uncorrected {H}amiltonian annealing.
\newblock In {\em Advances in Neural Information Processing Systems}, 2021.

\bibitem{geyer1991markov}
Charles~J Geyer.
\newblock Markov chain {M}onte {C}arlo maximum likelihood.
\newblock {\em Computing Science and Statistics, Proceedings of the 23rd
  Symposium on the Interface}, pages 156--163, 1991.

\bibitem{geyer1995annealing}
Charles~J Geyer and Elizabeth~A Thompson.
\newblock Annealing {M}arkov chain {M}onte {C}arlo with applications to
  ancestral inference.
\newblock {\em Journal of the American Statistical Association},
  90(431):909--920, 1995.

\bibitem{ghadimi_nonconvex_sgd_2013}
Saeed Ghadimi and Guanghui Lan.
\newblock Stochastic first- and zeroth-order methods for nonconvex stochastic
  programming.
\newblock {\em SIAM Journal on Optimization}, 23(4), 2013.

\bibitem{glynn_likelihood_1990}
Peter~W. Glynn.
\newblock Likelihood ratio gradient estimation for stochastic systems.
\newblock {\em Communications of the ACM}, 33(10):75--84, 1990.

\bibitem{gower_variance_2020}
Robert Gower, Mark Schmidt, Francis Bach, and Peter Richt\'{a}rik.
\newblock Variance-reduced methods for machine learning.
\newblock {\em Proceedings of the IEEE}, 108(11):1968--1983, 2020.

\bibitem{grenioux2023sampling}
Louis Grenioux, Alain Durmus, {\'E}ric Moulines, and Marylou Gabri{\'e}.
\newblock On sampling with approximate transport maps.
\newblock {\em arXiv:2302.04763}, 2023.

\bibitem{guo_overview_2023}
Tian-De Guo, Yan Liu, and Cong-Ying Han.
\newblock An overview of stochastic quasi-{Newton} methods for large-scale
  machine learning.
\newblock {\em Journal of the Operations Research Society of China},
  11(2):245--275, 2023.

\bibitem{gutmann2016bayesian}
Michael~U Gutmann, Jukka Cor, et~al.
\newblock {B}ayesian optimization for likelihood-free inference of
  simulator-based statistical models.
\newblock {\em Journal of Machine Learning Research}, 17(125):1--47, 2016.

\bibitem{haario_adaptive_2001}
Heiko Haario, Eero Saksman, and Joanna Tamminen.
\newblock An adaptive {M}etropolis algorithm.
\newblock {\em Bernoulli}, 7(2):223--242, 2001.

\bibitem{hagemann2022snf}
Paul Hagemann, Johannes Hertrich, and Gabriele Steidl.
\newblock Stochastic normalizing flows for inverse problems: a {M}arkov chains
  viewpoint.
\newblock {\em SIAM/ASA Journal on Uncertainty Quantification},
  10(3):1162--1190, 2022.

\bibitem{hoffman2019neutra}
Matthew Hoffman, Pavel Sountsov, Joshua~V Dillon, Ian Langmore, Dustin Tran,
  and Srinivas Vasudevan.
\newblock Neu{T}ra-lizing bad geometry in {H}amiltonian {M}onte {C}arlo using
  neural transport.
\newblock {\em arXiv:1903.03704}, 2019.

\bibitem{huggins2016coresets}
Jonathan Huggins, Trevor Campbell, and Tamara Broderick.
\newblock Coresets for scalable {B}ayesian logistic regression.
\newblock {\em Advances in Neural Information Processing Systems}, 29, 2016.

\bibitem{hukushima1996exchange}
Koji Hukushima and Koji Nemoto.
\newblock Exchange {M}onte {C}arlo method and application to spin glass
  simulations.
\newblock {\em Journal of the Physical Society of Japan}, 65(6):1604--1608,
  1996.

\bibitem{ivgi_dog_2023}
Maor Ivgi, Oliver Hinder, and Yair Carmon.
\newblock {DoG} is {SGD}'s best friend: {A} parameter-free dynamic step size
  schedule, 2023.
\newblock arXiv:2302.12022.

\bibitem{Jankowiak21}
Martin Jankowiak and Du~Phan.
\newblock Surrogate likelihoods for variational annealed importance sampling.
\newblock {\em arXiv:2112.12194}, 2021.

\bibitem{jarvenpaa2019efficient}
Marko J{\"a}rvenp{\"a}{\"a}, Michael~U Gutmann, Arijus Pleska, Aki Vehtari, and
  Pekka Marttinen.
\newblock Efficient acquisition rules for model-based approximate {B}ayesian
  computation.
\newblock {\em Bayesian Analysis}, 14(2):595--622, 2019.

\bibitem{kandasamy2017query}
Kirthevasan Kandasamy, Jeff Schneider, and Barnab{\'a}s P{\'o}czos.
\newblock Query efficient posterior estimation in scientific experiments via
  {B}ayesian active learning.
\newblock {\em Artificial Intelligence}, 243:45--56, 2017.

\bibitem{kim2022markov}
Kyurae Kim, Jisu Oh, Jacob Gardner, Adji~Bousso Dieng, and Hongseok Kim.
\newblock Markov chain score ascent: A unifying framework of variational
  inference with {M}arkovian gradients.
\newblock {\em Advances in Neural Information Processing Systems},
  35:34802--34816, 2022.

\bibitem{kim_guide_2015}
Sujin Kim, Raghu Pasupathy, and Shane Henderson.
\newblock A guide to sample average approximation.
\newblock In {\em Handbook of Simulation Optimization}, volume 216, pages
  207--243. Springer, 2015.

\bibitem{kingma_adam_2015}
Diederik~P. Kingma and Jimmy Ba.
\newblock Adam: {A} method for stochastic optimization.
\newblock In {\em {International} {Conference} on {Learning}
  {Representations}}, 2015.

\bibitem{kingma2014auto}
Diederik~P Kingma and Max Welling.
\newblock Auto-encoding variational {B}ayes.
\newblock In {\em International Conference on Learning Representations}, 2014.

\bibitem{kucukelbir2017automatic}
Alp Kucukelbir, Dustin Tran, Rajesh Ranganath, Andrew Gelman, and David~M Blei.
\newblock Automatic differentiation variational inference.
\newblock {\em Journal of Machine Learning Research}, 2017.

\bibitem{lei_nonconvex_sgd_2020}
Yunwen Lei, Ting Hu, Guiying Li, and Ke~Tang.
\newblock Stochastic gradient descent for nonconvex learning without bounded
  gradient assumptions.
\newblock {\em IEEE Transactions on Neural Networks and Learning Systems},
  31(10):4394--4400, 2020.

\bibitem{liu_limited_1989}
Dong~C. Liu and Jorge Nocedal.
\newblock On the limited memory {BFGS} method for large scale optimization.
\newblock {\em Mathematical Programming}, 45(1):503--528, 1989.

\bibitem{mahadevan_average_1996}
Sridhar Mahadevan.
\newblock Average reward reinforcement learning: {Foundations}, algorithms, and
  empirical results.
\newblock {\em Machine Learning}, 22(1):159--195, 1996.

\bibitem{maire_adaptive_2019}
Florian Maire, Nial Friel, Antonietta Mira, and Adrian~E. Raftery.
\newblock Adaptive incremental mixture {M}arkov chain {M}onte {C}arlo.
\newblock {\em Journal of Computational and Graphical Statistics},
  28(4):790--805, 2019.

\bibitem{manousakas2020bayesian}
Dionysis Manousakas, Zuheng Xu, Cecilia Mascolo, and Trevor Campbell.
\newblock Bayesian pseudocoresets.
\newblock {\em Advances in Neural Information Processing Systems},
  33:14950--14960, 2020.

\bibitem{marzouk2016introduction}
Youssef Marzouk, Tarek Moselhy, Matthew Parno, and Alessio Spantini.
\newblock An introduction to sampling via measure transport.
\newblock {\em arXiv:1602.05023}, 2016.

\bibitem{matthews2022craft}
Alex Matthews, Michael Arbel, Danilo~Jimenez Rezende, and Arnaud Doucet.
\newblock Continual repeated annealed flow transport {M}onte {C}arlo.
\newblock In {\em International Conference on Machine Learning}, pages
  15196--15219. PMLR, 2022.

\bibitem{meeds2014gps}
Edward Meeds and Max Welling.
\newblock {GPS}-{ABC}: {G}aussian process surrogate approximate {B}ayesian
  computation.
\newblock In {\em Uncertainty in Artificial Intelligence}, pages 593--602,
  2014.

\bibitem{meng_warp_2002}
Xiao-Li Meng and Stephen Schilling.
\newblock Warp {Bridge} {Sampling}.
\newblock {\em Journal of Computational and Graphical Statistics},
  11(3):552--586, 2002.

\bibitem{mohamed_monte_2020}
Shakir Mohamed, Mihaela Rosca, Michael Figurnov, and Andriy Mnih.
\newblock Monte {C}arlo gradient estimation in machine learning.
\newblock {\em Journal of Machine Learning Research}, 21(132):1--62, 2020.

\bibitem{montavon_neural_2012}
Grégoire Montavon, Geneviève~B. Orr, and Klaus-Robert Müller, editors.
\newblock {\em Neural {Networks}: {Tricks} of the {Trade}: {Second} {Edition}},
  volume 7700 of {\em Lecture {Notes} in {Computer} {Science}}.
\newblock Springer Berlin Heidelberg, 2012.

\bibitem{morgenstern_how_1985}
Jacques Morgenstern.
\newblock How to compute fast a function and all its derivatives: a variation
  on the theorem of {Baur}-{S}trassen.
\newblock {\em ACM SIGACT News}, 16(4):60--62, 1985.

\bibitem{moulines_non-asymptotic_2011}
{\'{E}}ric Moulines and Francis Bach.
\newblock Non-asymptotic analysis of stochastic approximation algorithms for
  machine learning.
\newblock In {\em Advances in {Neural} {Information} {Processing} {Systems}},
  volume~24, 2011.

\bibitem{murray2010elliptical}
Iain Murray, Ryan Adams, and David MacKay.
\newblock Elliptical slice sampling.
\newblock In {\em International Conference on Artificial Intelligence and
  Statistics}, pages 541--548. JMLR Workshop and Conference Proceedings, 2010.

\bibitem{naesseth2020markovian}
Christian Naesseth, Fredrik Lindsten, and David Blei.
\newblock Markovian score climbing: Variational inference with
  {KL}(p{\textbar}{\textbar}q).
\newblock {\em Advances in Neural Information Processing Systems},
  33:15499--15510, 2020.

\bibitem{naik2022fast}
Cian Naik, Judith Rousseau, and Trevor Campbell.
\newblock Fast {B}ayesian coresets via subsampling and quasi-{N}ewton
  refinement.
\newblock {\em Advances in Neural Information Processing Systems}, 35:70--83,
  2022.

\bibitem{nash_survey_2000}
Stephen~G. Nash.
\newblock A survey of truncated-{Newton} methods.
\newblock {\em Journal of Computational and Applied Mathematics},
  124(1):45--59, 2000.

\bibitem{neal_bayesian_1992}
Radford Neal.
\newblock Bayesian learning via stochastic dynamics.
\newblock In {\em Advances in {Neural} {Information} {Processing} {Systems}},
  volume~5, 1992.

\bibitem{nemirovski_robust_2009}
A.~Nemirovski, A.~Juditsky, G.~Lan, and A.~Shapiro.
\newblock Robust stochastic approximation approach to stochastic programming.
\newblock {\em SIAM Journal on Optimization}, 19(4):1574--1609, January 2009.

\bibitem{nemirovskii_problem_1983}
A.~S. Nemirovski and D.~B. Yudin.
\newblock {\em Problem Complexity and Method Efficiency in Optimization}.
\newblock Wiley-{Interscience} Series in Discrete Mathematics. Wiley,
  Chichester ; New York, 1983.

\bibitem{nesterov_method_1983}
Y.~Nesterov.
\newblock A method for solving the convex programming problem with convergence
  rate {$O(1/k^2)$}.
\newblock {\em Proceedings of the USSR Academy of Sciences}, 1983.

\bibitem{papamakarios2021normalizing}
George Papamakarios, Eric Nalisnick, Danilo~Jimenez Rezende, Shakir Mohamed,
  and Balaji Lakshminarayanan.
\newblock Normalizing flows for probabilistic modeling and inference.
\newblock {\em The Journal of Machine Learning Research}, 22(1):2617--2680,
  2021.

\bibitem{papamakarios2019sequential}
George Papamakarios, David Sterratt, and Iain Murray.
\newblock Sequential neural likelihood: Fast likelihood-free inference with
  autoregressive flows.
\newblock In {\em Artificial Intelligence and Statistics}, pages 837--848,
  2019.

\bibitem{paquet_perturbation_2009}
Ulrich Paquet, Ole Winther, and Manfred Opper.
\newblock Perturbation corrections in approximate inference: {M}ixture
  modelling applications.
\newblock {\em Journal of Machine Learning Research}, 10(43):1263--1304, 2009.

\bibitem{parno2018transport}
Matthew~D Parno and Youssef~M Marzouk.
\newblock Transport map accelerated {M}arkov chain {M}onte {C}arlo.
\newblock {\em SIAM/ASA Journal on Uncertainty Quantification}, 6(2):645--682,
  2018.

\bibitem{polyak64}
Boris Polyak.
\newblock Some methods of speeding up the convergence of iteration methods.
\newblock {\em USSR Computational Mathematics and Mathematical Physics},
  4(5):1--17, 1964.

\bibitem{polyak_new_1990}
Boris Polyak.
\newblock New stochastic approximation type procedures.
\newblock {\em Avtomatica i Telemekhanika}, 7:98--107, 1990.

\bibitem{price2018bayesian}
Leah~F Price, Christopher~C Drovandi, Anthony Lee, and David~J Nott.
\newblock {B}ayesian synthetic likelihood.
\newblock {\em Journal of Computational and Graphical Statistics}, 27(1):1--11,
  2018.

\bibitem{rakhlin_strcvx_2012}
Alexander Rakhlin, Ohad Shamir, and Karthik Sridharan.
\newblock Making gradient descent optimal for strongly convex stochastic
  optimization.
\newblock In {\em International Conference on Machine Learning}, 2012.

\bibitem{ranganath2014black}
Rajesh Ranganath, Sean Gerrish, and David Blei.
\newblock Black box variational inference.
\newblock In {\em Artificial Intelligence and Statistics}, pages 814--822.
  PMLR, 2014.

\bibitem{rasmussen2003gaussian}
Carl Rasmussen.
\newblock Gaussian processes to speed up hybrid {M}onte {C}arlo for expensive
  {B}ayesian integrals.
\newblock In {\em Bayesian Statistics}, pages 651--659, 2003.

\bibitem{j.2018on}
Sashank~J. Reddi, Satyen Kale, and Sanjiv Kumar.
\newblock On the convergence of {Adam} and beyond.
\newblock In {\em International Conference on Learning Representations}, 2018.

\bibitem{reddi2019convergence}
Sashank~J Reddi, Satyen Kale, and Sanjiv Kumar.
\newblock On the convergence of {A}dam and beyond.
\newblock {\em arXiv:1904.09237}, 2019.

\bibitem{rezende_flows_2015}
Danilo Rezende and Shakir Mohamed.
\newblock Variational inference with normalizing flows.
\newblock In {\em International Conference on Machine Learning}, 2015.

\bibitem{rezende2014stochastic}
Danilo~Jimenez Rezende, Shakir Mohamed, and Daan Wierstra.
\newblock Stochastic backpropagation and approximate inference in deep
  generative models.
\newblock In {\em International Conference on Machine Learning}, pages
  1278--1286, 2014.

\bibitem{robbins_monro_1951}
Herbert Robbins and Sutton Monro.
\newblock A stochastic approximation method.
\newblock {\em Annals of Mathematical Statistics}, 22(3):400--407, 1951.

\bibitem{roberts1997weak}
Gareth~O Roberts, Andrew Gelman, and Walter~R Gilks.
\newblock Weak convergence and optimal scaling of random walk {M}etropolis
  algorithms.
\newblock {\em The Annals of Applied Probability}, 7(1):110--120, 1997.

\bibitem{roberts1998optimal}
Gareth~O Roberts and Jeffrey~S Rosenthal.
\newblock Optimal scaling of discrete approximations to {L}angevin diffusions.
\newblock {\em Journal of the Royal Statistical Society: Series B (Statistical
  Methodology)}, 60(1):255--268, 1998.

\bibitem{roberts2009examples}
Gareth~O Roberts and Jeffrey~S Rosenthal.
\newblock Examples of adaptive {MCMC}.
\newblock {\em Journal of Computational and Graphical Statistics},
  18(2):349--367, 2009.

\bibitem{rubin1984bayesianly}
Donald~B Rubin.
\newblock Bayesianly justifiable and relevant frequency calculations for the
  applied statistician.
\newblock {\em The Annals of Statistics}, pages 1151--1172, 1984.

\bibitem{rubinstein_sensitivity_1992}
Reuven~Y. Rubinstein.
\newblock Sensitivity analysis of discrete event systems by the “push out”
  method.
\newblock {\em Annals of Operations Research}, 39(1):229--250, 1992.

\bibitem{rue_approximating_2004}
Håvard Rue, Ingelin Steinsland, and Sveinung Erland.
\newblock Approximating hidden {Gaussian} {Markov} random fields.
\newblock {\em Journal of the Royal Statistical Society: Series B (Statistical
  Methodology)}, 66(4):877--892, 2004.

\bibitem{ruppert_efficient_1988}
D.~Ruppert.
\newblock Efficient {Estimations} from a {Slowly} {Convergent}
  {Robbins}-{Monro} {Process}.
\newblock Technical report, Cornell University Operations Research and
  Industrial Engineering, 1988.

\bibitem{ruth_review_2024}
William Ruth.
\newblock A review of {Monte} {Carlo}-based versions of the {EM} algorithm,
  2024.
\newblock arXiv:2401.00945.

\bibitem{samsonov2022local}
Sergey Samsonov, Evgeny Lagutin, Marylou Gabri{\'e}, Alain Durmus, Alexey
  Naumov, and {\'{E}}ric Moulines.
\newblock Local-global {MCMC} kernels: {T}he best of both worlds.
\newblock {\em Advances in Neural Information Processing Systems},
  35:5178--5193, 2022.

\bibitem{sarkka2019applied}
Simo S{\"a}rkk{\"a} and Arno Solin.
\newblock {\em Applied Stochastic Differential Equations}, volume~10.
\newblock Cambridge University Press, 2019.

\bibitem{schmidt_minimizing_2017}
Mark Schmidt, Nicolas Le~Roux, and Francis Bach.
\newblock Minimizing finite sums with the stochastic average gradient.
\newblock {\em Mathematical Programming}, 162(1):83--112, 2017.

\bibitem{schulman_trust_2015}
John Schulman, Sergey Levine, Pieter Abbeel, Michael Jordan, and Philipp
  Moritz.
\newblock Trust region policy optimization.
\newblock In {\em {International} {Conference} on {Machine} {Learning}}, pages
  1889--1897. PMLR, 2015.

\bibitem{sun2018mcgd}
Tao Sun, Yuejiao Sun, and Wotao Yin.
\newblock On {M}arkov chain gradient descent.
\newblock {\em Advances in Neural Information Processing Systems}, 31, 2018.

\bibitem{surjanovic2023pigeons}
Nikola Surjanovic, Miguel Biron-Lattes, Paul Tiede, Saifuddin Syed, Trevor
  Campbell, and Alexandre Bouchard-C{\^o}t{\'e}.
\newblock Pigeons.jl: {D}istributed sampling from intractable distributions.
\newblock {\em arXiv:2308.09769}, 2023.

\bibitem{surjanovic2022parallel}
Nikola Surjanovic, Saifuddin Syed, Alexandre Bouchard-C{\^o}t{\'e}, and Trevor
  Campbell.
\newblock Parallel tempering with a variational reference.
\newblock {\em Advances in Neural Information Processing Systems}, 35:565--577,
  2022.

\bibitem{sutskever_importance_2013}
Ilya Sutskever, James Martens, George Dahl, and Geoffrey Hinton.
\newblock On the importance of initialization and momentum in deep learning.
\newblock In {\em {International} {Conference} on {Machine} {Learning}}, pages
  1139--1147, 2013.

\bibitem{swendsen1986replica}
Robert~H Swendsen and Jian-Sheng Wang.
\newblock Replica {M}onte {C}arlo simulation of spin-glasses.
\newblock {\em Physical Review Letters}, 57(21):2607, 1986.

\bibitem{syed2022nrpt}
Saifuddin Syed, Alexandre Bouchard-C{\^o}t{\'e}, George Deligiannidis, and
  Arnaud Doucet.
\newblock Non-reversible parallel tempering: {A} scalable highly parallel
  {MCMC} scheme.
\newblock {\em Journal of the Royal Statistical Society: Series B (Statistical
  Methodology)}, 84(2):321--350, 2022.

\bibitem{syed2021parallel}
Saifuddin Syed, Vittorio Romaniello, Trevor Campbell, and Alexandre
  Bouchard-C{\^o}t{\'e}.
\newblock Parallel tempering on optimized paths.
\newblock In {\em International Conference on Machine Learning}, pages
  10033--10042. PMLR, 2021.

\bibitem{tabak_flows_2013}
Esteban Tabak and Cristina Turner.
\newblock A family of nonparametric density estimation algorithms.
\newblock {\em Communications on Pure and Applied Mathematics}, 66(2):145--164,
  2013.

\bibitem{tabak_flows_2010}
Esteban Tabak and Eric Vanden-Eijnden.
\newblock Density estimation by dual ascent of the log-likelihood.
\newblock {\em Communications in Mathematical Sciences}, 8(1):217--233, 2010.

\bibitem{tataru2017statistical}
Paula Tataru, Maria Simonsen, Thomas Bataillon, and Asger Hobolth.
\newblock Statistical inference in the {W}right--{F}isher model using allele
  frequency data.
\newblock {\em Systematic Biology}, 66(1):e30--e46, 2017.

\bibitem{teh_consistency_2016}
YW~Teh, A~Thiery, and SJ~Vollmer.
\newblock Consistency and fluctuations for stochastic gradient {Langevin}
  dynamics.
\newblock {\em Journal of Machine Learning Research}, 17(7):1--33, 2016.
\newblock Publisher: Journal of Machine Learning Research.

\bibitem{welling_bayesian_2011}
Max Welling and Yee~Whye Teh.
\newblock Bayesian learning via stochastic gradient {L}angevin dynamics.
\newblock In {\em {International} {Conference} on {Machine} {Learning}}, pages
  681--688, 2011.

\bibitem{wilkinson2014accelerating}
Richard Wilkinson.
\newblock Accelerating {ABC} methods using {G}aussian processes.
\newblock In {\em Artificial Intelligence and Statistics}, pages 1015--1023,
  2014.

\bibitem{wood2010statistical}
Simon~N Wood.
\newblock Statistical inference for noisy nonlinear ecological dynamic systems.
\newblock {\em Nature}, 466(7310):1102--1104, 2010.

\bibitem{wu2020snf}
Hao Wu, Jonas K{\"o}hler, and Frank No{\'e}.
\newblock Stochastic normalizing flows.
\newblock {\em Advances in Neural Information Processing Systems},
  33:5933--5944, 2020.

\bibitem{younes_decreasing_ergodicity_1999}
Laurent Younes.
\newblock On the convergence of {M}arkovian stochastic algorithms with rapidly
  decreasing ergodicity rates.
\newblock {\em Stochastics and Stochastic Reports}, 65:177--228, 1999.

\bibitem{zhang2018advances}
Cheng Zhang, Judith B{\"u}tepage, Hedvig Kjellstr{\"o}m, and Stephan Mandt.
\newblock Advances in variational inference.
\newblock {\em IEEE Transactions on Pattern Analysis and Machine Intelligence},
  41(8):2008--2026, 2018.

\bibitem{Zhang21}
Guodong Zhang, Kyle Hsu, Jianing Li, Chelsea Finn, and Roger Grosse.
\newblock Differentiable annealed importance sampling and the perils of
  gradient noise.
\newblock In {\em Advances in Neural Information Processing Systems}, 2021.

\end{thebibliography}

\end{document}